\pdfoutput=1

\documentclass[11pt]{article}

\usepackage[final]{acl}

\usepackage{times}
\usepackage{latexsym}

\usepackage{multirow}
\usepackage{xcolor}
\usepackage{colortbl}

\usepackage{pifont}
\usepackage{subcaption}
\usepackage{listings}

\usepackage[T1]{fontenc}
\usepackage[utf8]{inputenc}

\usepackage{microtype}

\usepackage{inconsolata}
\usepackage{enumitem}
\usepackage{graphicx}
\usepackage{tabularx}
\usepackage{cleveref}
\usepackage{changepage,threeparttable}
\usepackage{booktabs}
\usepackage{graphicx}
\usepackage{float}

\usepackage{tikz}
\usepackage{pgfplots}
\usepackage{adjustbox}
\usepackage{hyperref}
\usepackage{xurl}

\crefname{section}{\S}{\S}

\crefname{figure}{Fig.}{Figures}

\crefname{table}{Tab.}{Tables}

\crefname{equation}{Eq.}{Equations}

\crefname{appendix}{Appendix}{Appendix}

\newcommand{\transparentmidrule}{\arrayrulecolor{white}\midrule\arrayrulecolor{black}}
\newcommand{\transparenttoprule}{\arrayrulecolor{white}\toprule\arrayrulecolor{black}}

\makeatletter
\newenvironment{chapquote}[2][2em]
  {\setlength{\@tempdima}{#1}%
   \def\chapquote@author{#2}%
   \parshape 1 \@tempdima \dimexpr\linewidth-2\@tempdima\relax%
   \itshape}
  {\par\normalfont\hfill--\ \chapquote@author\hspace*{\@tempdima}\par\bigskip}
\makeatother

\makeatletter
\DeclareRobustCommand\onedot{\futurelet\@let@token\@onedot}
\def\@onedot{\ifx\@let@token.\else.\null\fi\xspace}

\def\eg{\emph{e.g}\onedot}

 \def\vs{\emph{vs}\onedot}

\usepackage{xspace}
\usepackage{enumitem}
\usepackage{array}
\newcolumntype{C}{>{\centering\arraybackslash}X}
\usepackage{listings}
\usepackage{xcolor}
\usepackage[utf8]{inputenc}
\usepackage{tcolorbox}
\usepackage{fancyvrb}
\definecolor{codegreen}{rgb}{0,0.6,0}
\definecolor{codegray}{rgb}{0.5,0.5,0.5}
\definecolor{codepurple}{rgb}{0.58,0,0.82}
\definecolor{backcolour}{rgb}{0.95,0.95,0.92}

\newcommand{\bfsection}[1]{\noindent\textbf{#1}.}

\definecolor{salmon}{RGB}{250, 128, 114}

\newcommand\myfont[1]{\smash{{\usefont{T1}{qag}{m}{n}#1}}}

\newcommand{\modelnamefancy}{\myfont{VisArgs}\xspace}
\newcommand{\modelname}{VisArgs\xspace}

\title{Selective Vision is the Challenge for Visual Reasoning:\\A Benchmark for Visual Argument Understanding}

\newcommand*\samethanks[1][\value{footnote}]{\footnotemark[#1]}

\author{
Jiwan Chung$^{\spadesuit}$\thanks{~~denotes equal contribution} \quad
Sungjae Lee$^{\spadesuit}$\samethanks \quad
Minseo Kim$^{\spadesuit}$ \quad
Seungju Han$^{\diamondsuit\clubsuit}$ \\
\textbf{Ashkan Yousefpour}$^{\spadesuit}$ \quad
\textbf{Jack Hessel}$^{\heartsuit}$ \quad
\textbf{Youngjae Yu}$^{\spadesuit}$ \quad
\\
\small{$\spadesuit$ Yonsei University} \quad
\small{$\diamondsuit$ Seoul National University} \quad
\small{$\clubsuit$ Allen Institute for AI} \quad
\small{$\heartsuit$ Samaya AI} \quad \\
\texttt{jiwan.chung@yonsei.ac.kr}
}

\begin{document}
\maketitle

\begin{abstract}


Visual arguments, often used in advertising or social causes, rely on images to persuade viewers to do or believe something. Understanding these arguments requires selective vision: only specific visual stimuli within an image are relevant to the argument, and relevance can only be understood within the context of a broader argumentative structure.
While visual arguments are readily appreciated by human audiences, we ask: are today's AI capable of similar 
understanding?

We present \modelnamefancy\footnote{Data: \url{https://huggingface.co/datasets/jiwan-chung/visargs}}, a dataset of 1,611 images annotated with 5,112 visual premises (with regions), 5,574 commonsense premises, and reasoning trees connecting them into structured arguments. We propose three tasks for evaluating visual argument understanding: premise localization, premise identification, and conclusion deduction.
Experiments\footnote{Code: \url{https://github.com/JiwanChung/VisArgs}} show that 1) machines struggle to capture visual cues: GPT-4-O achieved 78.5\% accuracy, while humans reached 98.0\%. Models also performed 19.5\% worse when distinguishing between irrelevant objects within the image compared to external objects. 2) Providing relevant visual premises improved model performance significantly.

\end{abstract}



\section{Introduction}
\label{sec:intro}

\begin{chapquote}{\citet{lubbock1893beauties}}
What we see depends\\mainly on what we look for.
\end{chapquote}
\nocite{lubbock1893beauties}

Humans often communicate messages visually.
For example, traffic light colors regulate drivers' behavior, while computer icons, such as the trash bin symbol for deleting files or the magnifying glass for searching, guide user actions.

\begin{figure}
    \includegraphics[width=0.48\textwidth]{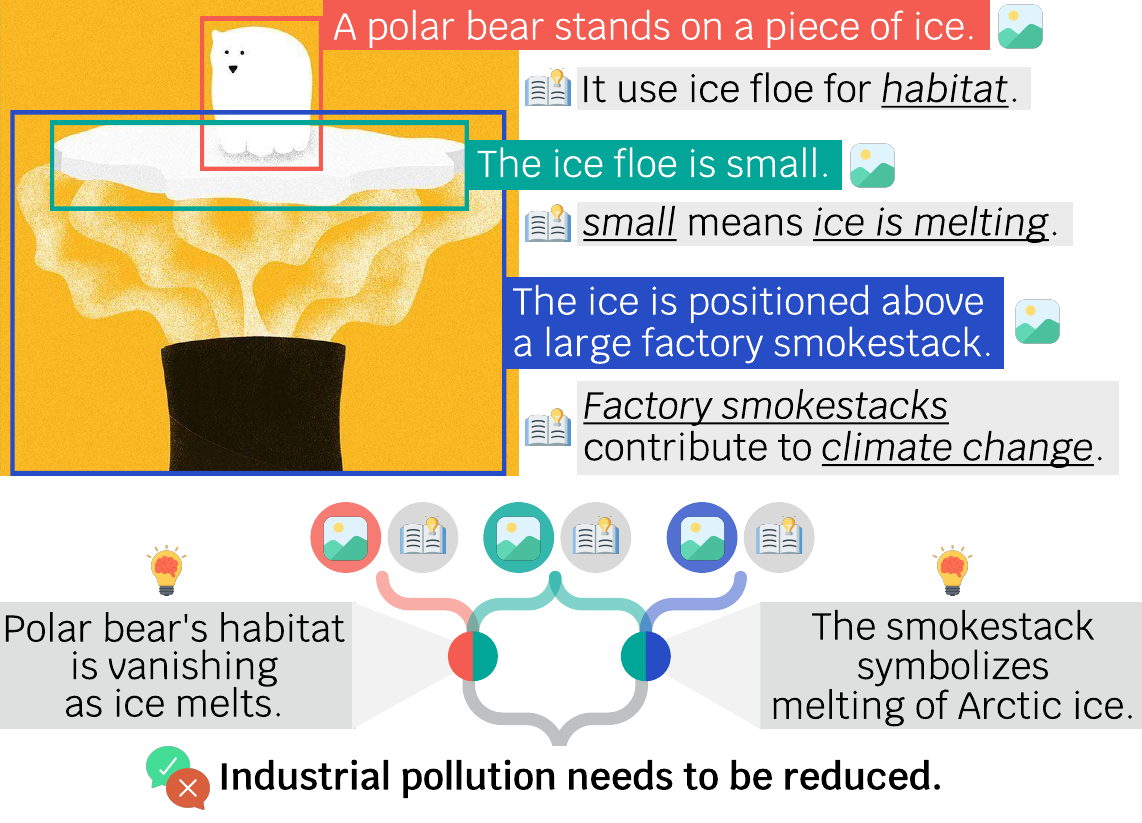}
    \caption{An example from our \modelnamefancy corpus. \modelname makes the persuasion process in a visual argument explicit by representing it as a reasoning tree. Image credit: Eglė Plytnikaitė}
    \label{fig:teaser}
\end{figure}

We consider the case of
\textit{visual arguments.} Consider \cref{fig:teaser}, which depicts a polar bear on a shrinking ice floe. Without any text, this image calls attention to climate change: a visual metaphor connects melting ice to industrial emissions from factories. A plausible interpretation of the argument concludes: \emph{industrial pollution needs to be reduced.} 

\begin{figure*}[!t]
    \includegraphics[width=1\textwidth]{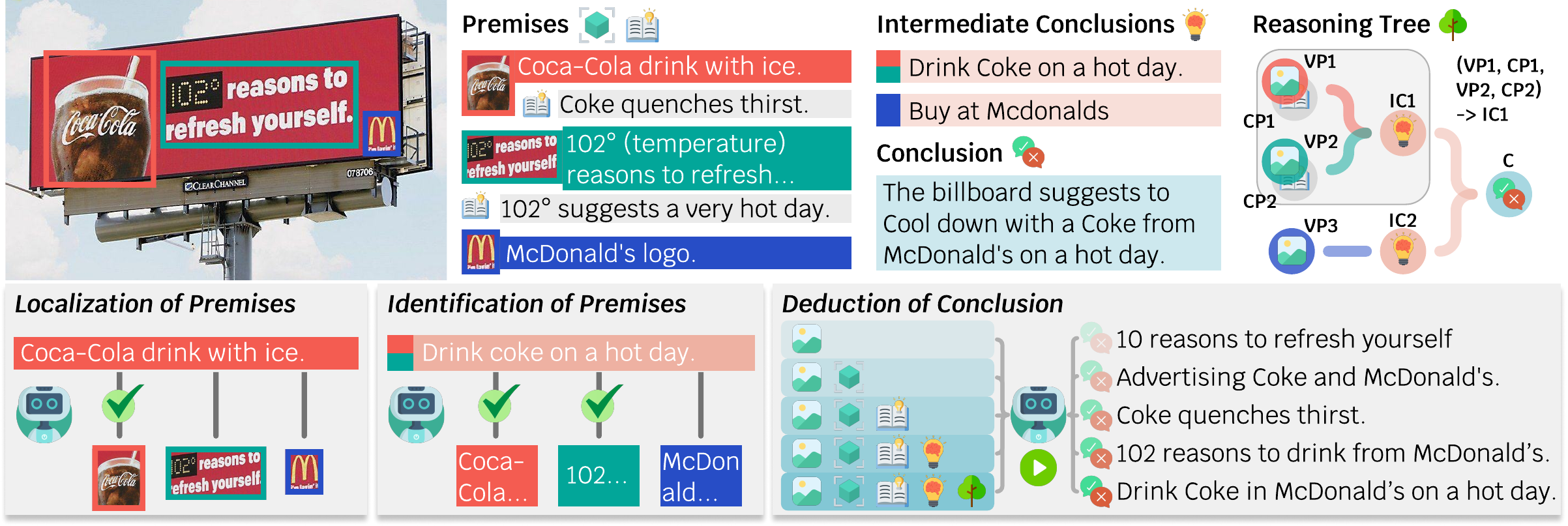}
    \caption{To identify the bottleneck in visual argument understanding, we define three tasks over \modelnamefancy: \textit{Localization of Premises} requires models to ground the visual premises. \textit{Identification of Premises} necessitates models to infer the visual premise relevant to the given intermediate conclusion. \textit{Deduction of Conclusion} studies the ability of models to deduce the argument's conclusion based on different levels of inputs.}
    \label{fig:fig2}
\end{figure*}



We introduce \modelnamefancy, an annotated dataset of 1,611 images containing visual arguments. 
\modelname makes explicit the reasoning process in interpreting a visual argument:\footnote{We note that our corpus contains just one possible \emph{interpretation} of a visual argument (rather than, e.g., claiming to represent the creator's intent).} each image is annotated with \textit{visual premises} grounded on object bounding boxes, \textit{commonsense premises} eliciting implicit knowledge, and \textit{argument trees}
formalizing the connection of these premises to the conclusion.
An argument tree consists of a root node (\textit{conclusion}), some internal nodes (\textit{intermediate conclusion}), and two types of leaf nodes (visual and commonsense premises).

Using \modelname, we propose three complementary tasks to evaluate different aspects of machine capacity for comprehending visual arguments as illustrated in~\cref{fig:fig2}: 1) \textit{Localization of Premises}: associates the description of a visual premise with a specific region in the image, 2) \textit{Identification of Premises}: Given an image and an (intermediate) conclusion, retrieves the necessary visual premises to support the conclusion, and 3)  \textit{Deduction of Conclusion}: generates the conclusion with increasing detail of the annotated visual argument.

Experiments on \modelname demonstrate that the main bottleneck for machine understanding of visual arguments is \emph{selective vision,} i.e., \textit{Identification of Premises} relevant to a given conclusion (see \cref{subsec:exp_id}). We show that while machines can identify visual premises within an image (albeit worse than human agreement, see \textit{Localization of premises} \cref{subsec:exp_loc}), they struggle to discern which premises are relevant to the conclusion among them. 
Results on our final \textit{Deduction of Conclusion} task (\cref{subsec:exp_con}) additionally support the hypothesis that difficulties in understanding visual arguments do not stem from 
deficiencies in
raw vision capacity. 
There, we controlled the level of input to the algorithm, ranging from raw images to explicit reasoning trees. 
The greatest accuracy gains came from the inclusion of \emph{relevant} visual cues, further supporting our main hypothesis. In all visual argument understanding tasks, machines perform worse than human agreement, providing avenues for future work.

In conclusion, our results suggest that selective attention to visual cues is the main bottleneck for the current AI capacity to understand visual arguments. This finding also establishes visual argument understanding as a distinct area of study in the computational domain: vision does not precede, but works jointly with reasoning in terms of understanding visual arguments. We expect that \modelnamefancy will be utilized as a diagnostic benchmark for selective vision in future multimodal models: even the best current models lag significantly behind human performance in our \textit{Identification of Premises} and \textit{Deduction of Conclusion} tasks.



\section{Related Work}
\label{sec:related_work}

\noindent\textbf{Visual arguments} are arguments built on visual medium~\cite{boland2005visual}.
Unlike typical images, a visual argument is intentionally organized to persuade viewers to a certain conclusion~\cite{birdsell1996toward,boland2005visual}. This work builds upon to ongoing debates in the human studies literature about the nature of visual arguments~\cite{johnson2003visual,tseronis2018multimodal}. Our results (\S~\ref{sec:exp}) suggest that understanding visual arguments
requires focusing on a subset of the visual context: not all visual cues contribute, and identifying the relevant ones is the key necessity.
This task one of \textit{selective vision}: the human capability to focus on behaviorally relevant stimuli.~\cite{desimone1995neural}.
Examples of visual arguments are prevalent in advertisements~\cite{kjeldsen2012pictorial,zhang2018equal,hussain2017automatic,ye2019interpreting}, cartoons~\cite{birdsell2007outlines}, mathematical educations~\cite{inglis2009persuasiveness}, and, arguably, diagrams \cite{kembhavi2016diagram,alikhani2018arrows}.
Liu et al.~\cite{liu2022imagearg} also investigate arguments conveyed through images. However, our work focuses on \textit{visual argument}, requiring that the argumentative content be primarily communicated through visual elements rather than relying solely on accompanying captions or written text. Furthermore, we provide explicit annotations of the argumentation structure in a tree format, facilitating detailed, hierarchical analysis of the model’s ability to comprehend visual arguments step-by-step.

\bfsection{Multimodal reasoning} Recent studies have introduced various multimodal models capable of sophisticated reasoning across different modalities, such as vision and language. Models such as LLaVA~\cite{liu2023llava}, Idefics2~\cite{laurenccon2024matters}, and Qwen-VL~\cite{Qwen-VL} are built on pretrained large language models (e.g., LLaMA~\cite{touvron2023llama}) and integrate vision encoders. Others, including OFA~\cite{wang2022ofa} and Unified-IO~\cite{lu2022unified}, are developed from scratch. These models excel in tasks such as localization, image captioning, and commonsense reasoning.
Furthermore, models such as Unified-IO-2~\cite{lu2023uio2} and GPT-4-O~\cite{achiam2023gpt} can understand audio, while others~\cite{zellers2022merlot, han2023champagne} support video understanding, demonstrating broad multimodal reasoning capabilities.

\bfsection{Beyond factual visual understanding}
Visual comprehension is moving beyond factual understanding to include various types of writing. 
These include visual commonsense reasoning~\cite{zellers2019vcr,park2020visualcomet, han2023reading, hessel2022abduction}, humor understanding~\cite{hessel2023androids, hyun2023smile}, and understanding social interaction~\cite{zadeh2018multimodal}. Of particular relevance to our work is visual metaphors~\cite{akula2023metaclue}, which express abstract concepts with concrete visual cues.
While some overlap exists in the images used, there are clear differences in intention and structure; not all metaphorical images present clear arguments and can be seen as visual arguments. Conversely, not all visual arguments depend on metaphors~\cite{blair2012possibility}. 

\bfsection{Argument structure}
An argument is typically understood as a structure that starts from a set of premises (reasons) and ends in a conclusion, often represented symbolically as a tree~\cite{whately1863elements,freeman2011dialectics}. While there have been extensions, including computational models of arguments~\cite{bench2007argumentation,rahwan2009argumentation,atkinson2017towards}, we use the basic form of trees connecting premises to conclusions, following previous literature~\cite{stab2014identifying,lawrence2020argument}.


\section{\modelnamefancy Dataset}
\label{sec:dataset}

\begin{figure}[!t]
    \includegraphics[width=0.48\textwidth]{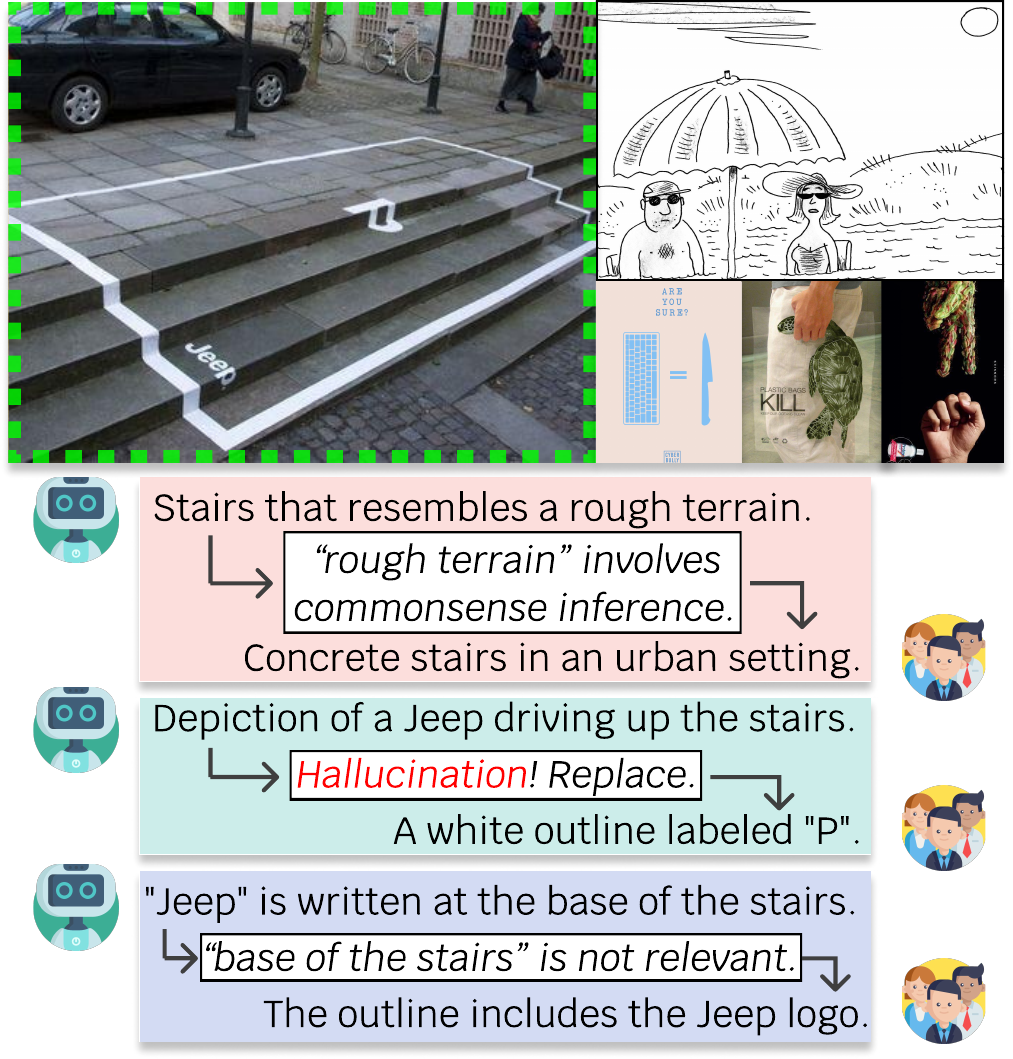}
    \caption{Human workers iteratively refine initial data produced by machines in \modelname annotation process.}
    \label{fig:annotation}
\end{figure}

\modelnamefancy comprises a total of 1,611 images featuring clear visual arguments. These images are categorized into 914 advertisement images and 697 cartoon images based on their sources. Each image in \modelname is annotated with descriptions and bounding boxes for the visual premises (\textsc{VP}), descriptions of the commonsense premises (\textsc{CP}), the conclusion, and an argumentation tree (T) detailing the reasoning path from the premises to the conclusion (C). All descriptions are in English, with an average character length of 79, 91, 142, and 105 for \textsc{VP}, \textsc{CP}, C, and T, respectively. On average, each image contains 3.17 visual premises, 3.46 commonsense premises, and 2.88 intermediate conclusions.

\subsection{Annotation Process}
\label{subsec:data_ann}

We partially rely on GPT-4-O~\cite{achiam2023gpt} for initial annotations. However, these machine-generated annotations serve only as preliminary seeds, which are then extensively refined by experienced human workers, as illustrated in~\S\ref{fig:annotation}. The machine's role is merely to provide imperfect starting points to facilitate the human annotation process. Below, we detail our annotation procedure.

\bfsection{Collecting Images}
Our primary criterion was to select images that enable human annotators to easily and accurately interpret both the visual premise and the corresponding conclusion, thereby clarifying the argumentative structure within the image.
Also, we ruled out samples with scene text within the images that directly describe the conclusion. 
We manually collect around 1,600 images following these criteria from Pinterest.\footnote{\url{www.pinterest.com}}
Starting with keyword-based searches (\eg \textit{creative ads}), we expanded our collection by exploring related images.
Cartoons (which often contain visual arguments~\cite{birdsell1996toward}) were sourced from a dedicated website.\footnote{\url{www.cartoonmovement.com}} We manually collected around 1,600 cartoons from various categories, including politics, education, and environment.
We include URLs to the images to comply with licensing terms following previous work~\cite{schuhmann2022laion,lee2021acav100m}.
Refer to Appendix~\ref{sec:ax_data} for details.
\bfsection{Describing Visual Premises}
The next step is to explicitly describe the visual argument within each image. However, during the early stages of our annotation process, we discovered that although humans can naturally understand visual arguments, they often find it challenging to articulate their interpretation into structured argumentation trees.
Therefore, we used an AI model (GPT-4-O) to generate initial candidates. Human workers then select and modify these initial annotations, as shown in~\cref{fig:annotation}. 
The human annotators could optionally incorporate new visual premises when necessary: $\sim21\%$ of images had their set of visual premises expanded through this process.
To facilitate this process, we break down the annotation into two steps: describing the visual premises and specifying the argument structure.

Given an image containing a visual argument, we instructed the model to generate a set of visual premises necessary to support the argument (refer to~\cref{sec:ax_data_prompt} for further details).
However, the AI model often fails to fully comprehend the visual argument. 
To address this, we engaged a pool of experienced human workers to review the machine-generated outputs. They selected the correct visual premises and made necessary modifications to ensure accuracy and coherence.
Additionally, we identified that a model-generated visual premise sometimes contains multiple atomic premises. We instructed the reviewers to separate these merged premises into individual atomic premises.
Further details are provided in~\cref{sec:ax_data}.


\bfsection{Specifying Argument Structure}
Given the visual premises and the image, we further annotate three components constituting the argumentation structure: commonsense premises, conclusions, and argument trees. As in the previous stage, we first generate initial candidates using an AI model.
For this stage, we impose an additional criterion: the set of selected premises should be both necessary and complete (refer to~\cref{sec:ax_data_prompt}). The same pool of human workers then adjust the annotations for greater accuracy.
The workers first verify the correctness of the conclusion and discard the image if it is incorrect. They then identify and correct any errors, including semantic and structural mistakes. We discarded 1,593 of the 3,204 images in this process. Details are provided in~\cref{sec:ax_data}.



\bfsection{Visual Grounding}
Lastly, we manually gather bounding box annotations for each visual premise to finalize the multimodal annotations. We assume a one-to-one relationship between each bounding box ($vp^r_i$) and its corresponding textual description ($vp^d_i$). Annotators are instructed to ensure accurate matching and precise bounding box tightness, as detailed in~\cref{sec:ax_data}. 


\subsection{Data Analysis}
\label{subsec:data_exp}

\begin{figure}
    \includegraphics[width=0.48\textwidth]{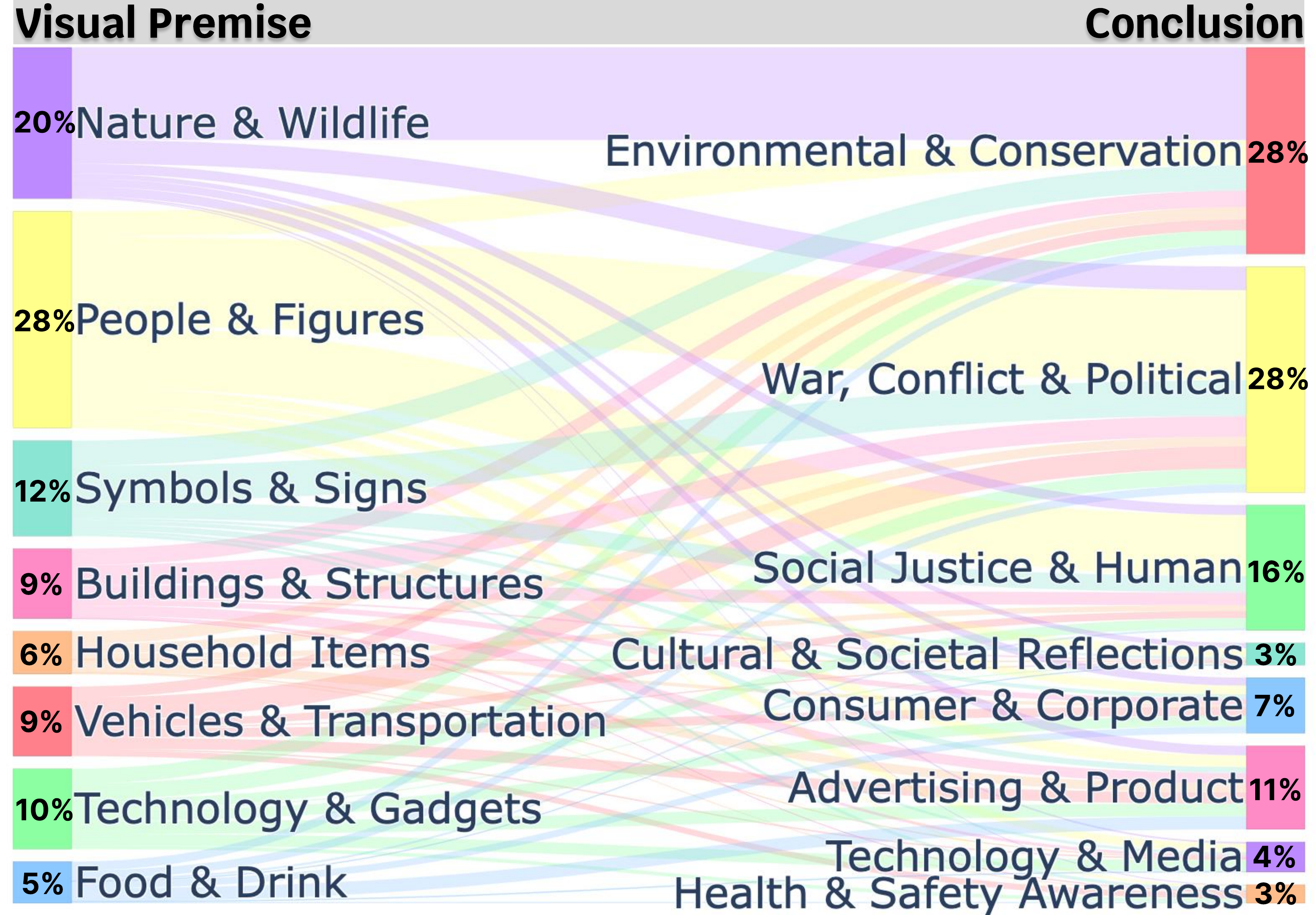}
    \caption{Variety of the topics represented in the visual premises and conclusions in \modelname.}
    \label{fig:vp_cp_class}
\end{figure}

\bfsection{Topic Diversity}
To gauge the diversity of topics covered in \modelname, we run zero-shot categorization using GPT-4-O and LLaMa3~\cite{llama3modelcard} to classify the topics of visual premises and conclusions.
The topics cover a wide range of visual objects and argument topics, as shown in~\cref{fig:vp_cp_class}.
Refer to~\cref{sec:ax_topic} for details.

\begin{table}[t]
    \centering
    \small
    \begin{tabular}{l|cc|cc}
    \toprule
    & \multicolumn{2}{c|}{Number of Samples} & \multicolumn{2}{c}{Image Size (pixels)} \\
    Category & Total & W/ Text & Width & Height \\
    \midrule
    Ads &  914& 389& 877.2 & 969.7 \\
    Cartoons &  697& 218& 480.0 & 427.6 \\
    \bottomrule
    \end{tabular}
    \caption{Overview of dataset statistics. \textit{W/ Text} indicates the subset of images containing scene text.}
    \label{tab:data_stats}
\end{table}

\begin{table}[t]
    \centering
    \small
    \begin{tabular}{l|cc}
    \toprule
    & Recall & Hit rate \\
    \midrule
    LLaVaNeXT &0.48  &0.14 \\
    LLaVa-LLaMa3-Docci &0.27  &0.02 \\
    ShareCaptioner &0.40  &0.12 \\
    \bottomrule
    \end{tabular}
    \caption{Frequency of detailed captions containing visual premises. \textit{Hit rate} denotes how often all visual premises per image are included in the captions.}
    \label{tab:detailed_captions}
\end{table}

\bfsection{Visual Cues \vs Dense Captioning}
In theory, selective attention to visual premises could be collapsed into an NLP problem by describing \textit{everything} in an image. To test this counter-hypothesis, we manually check how often the visual premises are contained in the outputs of detailed captioning models. We include three baselines here: a generalist (LLaVA-Next~\cite{liu2024llavanext}), a specialist (ShareCaptioner~\cite{chen2023sharegpt4v}), and LLaVA-LLaMa3~\cite{2023xtuner} fine-tuned on a detailed captioning corpus (DOCCI~\cite{OnoeDocci2024})\footnote{\url{huggingface.co/gokaygokay/llava-llama3-docci}}.
\cref{tab:detailed_captions} summarizes our manual inspection of 100 images, showing that the detailed captions insufficiently capture the visual premises, with the hit rate staying below $15\%$ for all models.

\bfsection{Safety}
Since we did not initially filter for safety, we now analyze the safety of \modelname using standard models. For textual safety, we utilize the Perspective API\footnote{\url{www.perspectiveapi.com}; June 2024 version.}, and for visual domains, we employ LAION-Safety\footnote{\url{www.github.com/LAION-AI/LAION-SAFETY}}. The toxicity scores for textual descriptions were $0.03$ for visual premises and $0.07$ for conclusions. Also, given the threshold of $0.7$, no descriptions and visual premises were classified as toxic.
Furthermore, only $71$ among $1611$ images are classified as unsafe.
Manual inspection reveals that such ``unsafe" images were social campaigns advocating \emph{against} the harmful behaviors which presumably triggered the LAION detector.


\section{Task Overview}
\label{sec:task}

We pose three tasks based on \modelname for a structured analysis of how machines understand arguments presented in visual form.

\begin{table}[t]
\centering
    \centering
    \small
    \begin{tabular}{l|c|ccc|c}
    \toprule
    & Acc. & Prec. & Rec. & F1 & Corr. ($\rho$) \\
    \midrule
    BLEU-4 &67 & 44 & 67 & 53 & 18 \\
    ROUGE &75 & 76 & 75 & 72 & 35\\
    CIDEr &72 & 70 & 72 & 70 & 26 \\
    GPTEval &75 & 83 & 75 & 76 & 53 \\
    \textbf{BERTScore} &\textbf{94} & \textbf{94} & \textbf{93} & \textbf{93} & \textbf{59} \\
    \bottomrule
    \end{tabular}
    \caption{Correlation of each metric with human decisions in the \textit{Deduction of Conclusion} task.}
    \label{tab:score_comparison}
\end{table}

\textbf{An instance of \modelname} consists of an image $I$, a set of visual premises $\textsc{VP} = \{(vp^{d}_{0}, vp^{r}_{0}), (vp^{d}_{1}, vp^{r}_{1}), \ldots\}$ with textual description $vp^{d}$ along with region grounding with a bounding box $vp^{r} = \langle x,y,h,w \rangle$,
a set of commonsense premises $\textsc{CP} = \{cp^{d}_{0}, cp^{d}_{1}, \ldots \}$, and
the conclusion in textual form $\textsc{C}$.
Further, a single argument tree for each image is built on the premises.
Each tree $t \in T$ represents a reasoning path leading to the conclusion $\textsc{C}$.
The nodes $N$ of a tree consist of the following:
1) leaf nodes: subsets of the union of the visual and commonsense premises $\textsc{VP} \cup \textsc{CP}$.
2) internal nodes: elements of the set of intermediate conclusions $\textsc{IC}$.
3. root node: the conclusion $\textsc{C}$.
An edge $e$ of the tree connects a subset of nodes $\bar{N} \subset \textsc{VP} \cup \textsc{CP} \cup \textsc{IC}$ to either an intermediate conclusion $ic \in \textsc{IC}$ or the final conclusion $\textsc{C}$.

\subsection{Localization of Premises}
\label{subsec:task_vp}
The first task focuses on assessing whether machines can accurately align visual premises ($\textsc{VP}^d$) with the corresponding regions ($\textsc{VP}^r$) in a given image ($I$), requiring minimal computational reasoning capabilities. It aims to determine if difficulties in understanding visual arguments originate from basic object detection stages.

We investigate two setups based on the algorithm's ability to output bounding box labels:
First, \textit{closed-set grounding} is designed for a broad range of models that lack explicit grounding capabilities. The problem is formulated as a retrieval task where the goal is to match a region in the image ($vp^r_i$) with an appropriate description ($vp^d_i$). We adapt standard image-text matching models (\eg CLIP) to perform grounded image-text matching. More details can be found in~\cref{sec:exp}.
Second, \textit{open-set grounding} tests models with explicit grounding capabilities. The task is framed as a visual grounding problem~\cite{yu2016modeling}, where the machine must locate an object in an image based on a natural language expression. Both the ground truth and machine output are represented as bounding box coordinates $\langle x, y, h, w \rangle$. Performance is evaluated using the intersection over union (IoU) ratio, with predictions considered correct if IoU $\ge 0.5$. 

%

\subsection{Identification of Premises}
\label{subsec:task_id}

The second task tests the machines' capabilities to discern visual premises that would better support the given conclusion.
Given the image $I$, the intermediate conclusion $ic$, and a superset of the gold text descriptions of the visual premises $S \supset \textsc{VP}^d$, the machine should retrieve a correct visual premise $vp^d_i \in \textsc{VP}^d$.
Note that the candidate set $S$ contains a single ground truth premise $vp^d_i$ and a fixed number $K=2$ of negative premises.

The complexity of a retrieval task is impacted by the choice of the negative set.
We explore four types of \textit{global} samplers and a single \textit{local} sampler for constructing the negative set.
The global samplers source the negatives from visual premises that \textit{do not} correspond to the selected image.
The only difference is the sample selection strategy:
1. \textit{Random} sampling samples uniformly without replacement. 
2. \textit{Visual} sampling samples from the top premise descriptions that are the closest to the given image. We use CLIPScore~\cite{hessel2021clipscore} for the multimodal scoring.
3. \textit{Textual} sampling samples from the top premise descriptions that are the closest to the ground truth premise. We use cosine similarity on the ColBERT~\cite{khattab2020colbert} representation space for the textual scoring.
4. \textit{Mixed} sampling combines textual and visual sampling by visually selecting from the top 10 textual retrieval results.

For \textit{local} sampling, we select from the visual premises that \textit{do} correspond to the given image.
Relying on our argumentation tree annotation, we can automatically obtain the set of local visual premises that does not help justify the given intermediate conclusion $ic$. we sample uniformly without duplicates from the local pool and name the method 5. \textit{Semantic} sampling due to its argumentation-dependent nature.
Additionally, we report human performance on 100 random samples to mitigate the risk of false negatives.

\subsection{Deduction of Conclusion}
\label{subsec:task_conclusion}
The final task is to evaluate how each component ($I$, $\textsc{VP}$, $\textsc{CP}$, $\textsc{IC}$, and $T$) influences the deduction of the conclusion $\textsc{C}$. We approach this as a sequence-to-sequence task aimed at generating $\textsc{C}$. While this allows flexible output formats, it complicates evaluation because the machine-generated text must be compared to the free-form label.
Common text comparison practices, such as BLEU~\cite{papineni2002bleu}, ROUGE~\cite{lin2004rouge}, and CIDER~\cite{vedantam2015cider} measure surface form similarity, not semantic similarity between conclusions. Alternatively, prompt-based evaluation using general reasoners (\eg GPT-4)~\cite{achiam2023gpt} can be biased by factors including candidate order~\cite{pezeshkpour2023large}. Human verification, though ideal, is costly and hard to reproduce.
We conduct a small-scale comparison study (see~\cref{tab:score_comparison}) to verify that the model-based metric BERTScore~\cite{Zhang*2020BERTScore} provides the most stable estimate, making it our primary metric. Details are in~\cref{sec:ax_score_comparison}.

\begin{table}[t]
    \small
    \centering
    \begin{tabular}{l|ccc}
    \toprule
    & \multicolumn{3}{c}{Acc. ($\%$)} \\
    \midrule
    & Ads & Cartoon & All \\
    \midrule
    Random & 33.33 & 33.33 & 33.33 \\
    Human & 100.00 & 100.00 & 100.00 \\
    \midrule
    CLIP\textsubscript{RN50} & 80.83 & 82.72 & 81.91 \\
    CLIP\textsubscript{ViT-L} & 82.72 & 82.96 & 82.85 \\
    CLIP\textsubscript{ViT-L@336} & 82.09 & 83.26 & 82.76 \\
    SigLIP & 86.10 & 86.67 & 86.43 \\
    AlphaCLIP & 75.15 & 77.44 & 76.45 \\
    \midrule
    OFA\textsubscript{Base} & 68.75 & 75.71 & 72.71  \\
    OFA\textsubscript{Large} & 72.01 & 79.18 & 76.10 \\ 
    \bottomrule
    \end{tabular}
    \captionof{table}{\textit{closed-set} results in \textit{localization of premises}.}
    \label{tab:task1_closed}
    \centering
    \small
    \begin{tabular}{l|cc}
    \toprule
    & IoU & Acc. ($\%$) \\
    \midrule
    UNINEXT-H &38.75 & 35.58 \\
    LISA & 44.25 & 44.62 \\
    Unified-IO-2 &48.61 & 47.15 \\
    OFA &50.14 & 49.13 \\
    MM-G-Dino &55.02 & 54.98 \\
    \bottomrule
    \end{tabular}
    \caption{\textit{open-set} results in \textit{localization of premises}.}
    \label{tab:task1_open}
\end{table}

\begin{table*}[t]
    \small
    \centering
    \begin{minipage}[t]{0.65\textwidth}
        \centering
        \begin{tabular}{l|cccc|l}
        \toprule
        & \multicolumn{4}{c|}{Global} & \multicolumn{1}{c}{Local} \\
        \midrule
        & Random & Visual & Textual & Mixed & \multicolumn{1}{c}{Semantic} \\
        \midrule
        Random & 33.33 & 33.33 & 33.33 & 33.33 & \multicolumn{1}{c}{33.33 (-)} \\
        Human & 100.00 & 99.00 & 94.00 & 100.00 & 98.00 ($\uparrow$ 4.00) \\
        \midrule
        OFA & 0.00 & 0.00 & 0.00 & 0.00 & \multicolumn{1}{c}{0.00 (-)} \\
        Qwen-VL-Chat & 86.05 & 85.77 & 70.67 & 75.57 & 49.74 ($\downarrow$ \textbf{20.93}) \\
        CogVLM & 97.46 & 96.39 & 88.00 & 92.22 & 65.31 ($\downarrow$ \textbf{22.69}) \\
        Idefics2 & 98.68 & 97.83 & 91.80 & 95.07 & 75.01 ($\downarrow$ \textbf{16.79}) \\
        InstructBLIP & 83.77 & 79.23 & 66.95 & 71.37 & 61.90 ($\downarrow$ \textbf{5.05}) \\
        Unified-IO-2 & 98.42 & 96.99 & 86.87 & 92.81 & 34.74 ($\downarrow$ \textbf{52.13}) \\
        LLaVA-1.5 & 98.65 & 97.91 & 83.74 & 89.86 & 67.43 ($\downarrow$ \textbf{16.31}) \\
        LLaVA-NeXT & 97.66 & 96.20 & 80.90 & 85.86 & 78.53 ($\downarrow$ \textbf{2.37}) \\ 
        GPT-4-O &  - & - & - & - & \multicolumn{1}{c}{79.50 (-)} \\
        \bottomrule
        \end{tabular}
    \end{minipage}
    \hspace{0.01\textwidth}
    \begin{minipage}[t]{0.15\textwidth}
        \raggedright
        \begin{tabular}{l}
            \transparenttoprule
            \\
            \transparentmidrule
            \multicolumn{1}{c}{Semantic} \\
            \midrule
            \\
            \multicolumn{1}{c}{+ G.T region} \\
            \midrule
            \multicolumn{1}{c}{-}\\
            \multicolumn{1}{c}{-}\\
            \multicolumn{1}{c}{-}\\
            \multicolumn{1}{c}{-}\\
            78.13 ($\uparrow$ 16.23) \\
            84.39 ($\uparrow$ 49.65) \\
            76.67 ($\uparrow$ 9.24) \\
            82.19 ($\uparrow$ 3.66) \\
            \multicolumn{1}{c}{-}\\
        \bottomrule
        \end{tabular}
    \end{minipage}
    \captionof{table}{Results of the \textit{Identification of Premises} task. Difference between the lowest score in \textit{global} and \textit{local} setup for each model are highlighted.}
    \label{tab:task2}
\end{table*}

\section{Experiments}
\label{sec:exp}


\subsection{Localization of Premises}
\label{subsec:exp_loc}

\textit{Localization of Premises} tests the visual grounding capabilities of machines. Given the image $I$ and description of a visual premise $vp^d$, the goal is to find a corresponding region $vp^r$ in the image.

\bfsection{Metrics and Models}
We define \textit{open-set} evaluation as a setting in which models are required to generate bounding box coordinates without relying on a predefined candidate list. As a result, models used for \textit{open-set} and \textit{closed-set} evaluations are architecturally distinct, since models lacking an explicit generative head, such as CLIP~\cite{radford2021learning}, are not compatible with open-set evaluation due to their dependence on a candidate region list for text-to-region matching.

For \textit{closed-set grounding}, which is an N-way classification task, the goal is to match the given description with the correct bounding box. To evaluate standard image-text matching algorithms (\eg CLIP), we crop the regions accordingly. The models for this task include various CLIP-based models (CLIP~\cite{radford2021learning} with different backbones and SigLIP~\cite{zhai2023sigmoid}) and a multitask model OFA~\cite{wang2022ofa}.
The CLIP-based models are adapted as follows: for each candidate object region (specified by bounding box coordinates), the corresponding regions are cropped from the image to create region-level image representations. Image features for each cropped region are then extracted using a CLIP-based encoder, while text features are obtained by encoding the input description using the same model. Cosine similarity is calculated between the text feature and each region-level image feature. The region with the highest similarity score is selected as the predicted match.

For \textit{open-set grounding}, which is to locate an object in an image based on a natural language expression, we instruct the models to output bounding box coordinates and we compare them to the ground truth region. A predicted coordinate is considered correct if its intersection over union with the gold label is at least (IoU $\ge 0.5$). We use a diverse set of models that support local region output formats, UNINEXT-H~\cite{UNINEXT}, LISA~\cite{lai2023lisa}, Unified-IO-2~\cite{lu2023uio2}, OFA, MM-G-DINO~\cite{liu2023grounding}. 

\bfsection{Results}
\cref{tab:task1_closed} demonstrates that current models are generally effective in matching descriptions of visual premises to the correct regions in images, thereby meeting the basic vision requirements for understanding visual arguments. However, the results for \textit{open-set} grounding, shown in \cref{tab:task1_open}, are somewhat mixed: the scores are acceptable but not uniformly high. We traced this performance decline to the nature of zero-shot object detectors, which are designed to detect concrete objects and clear segments. In contrast, our bounding boxes are more semantic~\cite{guo2018review}. Visual examples can be found in~\cref{sec:ax_open_grounding_samples}.

\begin{table}[!t]
    \centering
    \small
    \begin{tabular}{
    l@{\hspace{0.3\tabcolsep}}
    r@{\hspace{0.4\tabcolsep}}
    r@{\hspace{0.4\tabcolsep}}
    r@{\hspace{0.4\tabcolsep}}
    r@{\hspace{0.4\tabcolsep}}}
    \toprule
    & Image & + VP & + CP & + Tree \\
    \midrule
    
    LLaMA3 & - & 30.2 & 37.8 (\scriptsize$\uparrow$\textbf{7.6}) & 40.8 (\scriptsize$\uparrow$2.0) \\
    Mistralv0.2 & - & 18.9 & 30.2 (\scriptsize$\uparrow$\textbf{11.3}) & 36.6 (\scriptsize$\uparrow$6.4) \\
    Zephyr & - & 20.6 & 28.7 (\scriptsize$\uparrow$\textbf{8.1}) & 36.5 (\scriptsize$\uparrow$7.8) \\
    \midrule
    OFA & -41.3 & -24.6 (\scriptsize$\uparrow$\textbf{16.7}) & -16.5 (\scriptsize$\uparrow$8.1) & -13.9 (\scriptsize$\uparrow$2.6) \\
    Qwen-VL-Chat & 12.8 & 23.7 (\scriptsize$\uparrow$\textbf{10.9}) & 30.2 (\scriptsize$\uparrow$6.5) & 32.7 (\scriptsize$\uparrow$2.5) \\
    CogVLM & 25.7 & 30.7 (\scriptsize$\uparrow$\textbf{5.0}) & 33.6 (\scriptsize$\uparrow$2.9) & 36.3 (\scriptsize$\uparrow$2.7) \\
    Idefics2 & 16.4 & 22.8 (\scriptsize$\uparrow$6.4) & 29.5 (\scriptsize$\uparrow$6.7) &  36.6 (\scriptsize$\uparrow$\textbf{7.2}) \\
    InstructBLIP & -18.4 & 16.6 (\scriptsize$\uparrow$\textbf{35.0}) & 28.9 (\scriptsize$\uparrow$12.3) & 32.2 (\scriptsize$\uparrow$3.3) \\
    Unified-IO-2 & -9.9 & -3.4 (\scriptsize$\uparrow$6.5) & 4.2 (\scriptsize$\uparrow$\textbf{7.6}) & 8.0 (\scriptsize$\uparrow$3.8) \\
    LLaVA-1.5 & 2.2 & 20.0 (\scriptsize$\uparrow$\textbf{17.8}) & 29.6 (\scriptsize$\uparrow$9.6) & 33.7 (\scriptsize$\uparrow$4.1) \\
    LLaVA-Next & 15.1 & 28.4 (\scriptsize$\uparrow$\textbf{13.3}) & 34.3 (\scriptsize$\uparrow$5.9) & 39.5 (\scriptsize$\uparrow$5.2) \\
    GPT-4-O & 25.5 & - & 34.3 (\scriptsize$\uparrow$\textbf{8.8}) & 41.0 (\scriptsize$\uparrow$6.7) \\
    \bottomrule
    \end{tabular}
    \caption{Results of the \textit{Deduction of Conclusion} task, showing how incremental additions of inputs affect the correctness of the conclusion. Scores are presented using BERTScore, with similar trends observed across other metrics as detailed in~\cref{sec:ax_full_results}.}
    \label{tab:task3_small}
\end{table}

\subsection{Identification of Premises}
\label{subsec:exp_id}

\textit{Identification of Premises} tests the selective attention capabilities, i.e., selecting necessary visual cues to understand an argument.
Given the image $I$ and an intermediate conclusion $ic$, the goal is to select a visual premise $vp^d$ that leads to this intermediate conclusion.

\bfsection{Metrics and Models}
For this task, we retain only intermediate conclusions that have at least two unrelated visual premises within the image. We report classification accuracy based on a single gold visual premise and two negative candidates. The negative sets are sourced as described in~\cref{subsec:task_id} and are categorized into \textit{random}, \textit{visual}, \textit{textual}, \textit{mixed}, and \textit{semantic} sets.
Given the task's requirement for understanding argumentation structure, the models evaluated are primarily multimodal large language models with adequate reasoning capabilities. We experiment with a broad selection of models: OFA~\cite{wang2022ofa},  Qwen-VL-Chat~\cite{Qwen-VL},  CogVLM~\cite{wang2023cogvlm}, Idefics2~\cite{laurenccon2024matters},  InstructBLIP~\cite{dai2024instructblip}, Unified-IO 2~\cite{lu2023uio2}, LLaVa-1.5~\cite{liu2024improvedllava}, and LLaVa-Next~\cite{liu2024llavanext}.
For the sake of brevity, we do not report per-category results (\textit{Ads} and \textit{Cartoon}) here. Refer to \cref{sec:ax_full_results} for full results.

\bfsection{Results}
\cref{tab:task2} highlights a significant trend: models struggle to distinguish negatives within the image (\textit{local}), but excel in identifying \textit{global} negatives.
A major challenge for most models was handling \textit{semantic} negatives within the same image, as evidenced by the generally wide margin between models' performance on \textit{global} and \textit{local} setups. 
Still, the \textit{global} negative samples exhibited more pronounced distinctions based on their sampling scheme.
Negatives sampled uniformly were distinguishable by most models with $\ge90\%$ accuracy. In contrast, retrieval methods proved more challenging across the board, particularly for negatives retrieved using the text-to-text similarity model (\textit{textual}), which increased the problem complexity for most models.
Notably, OFA failed to follow zero-shot instructions for multiple-choice answering, scoring close to zero. 
Finally, we also present results for cropped ground-truth region images. Although cropped images are not lossless representations of the regions, all models exhibited significant improvements, indicating that the ability to infer relevant visual cues is indeed a critical challenge.
Thus, we conclude that models struggle to infer which visual cues support the argument.

\begin{figure}[!t]
    \includegraphics[width=0.48\textwidth]{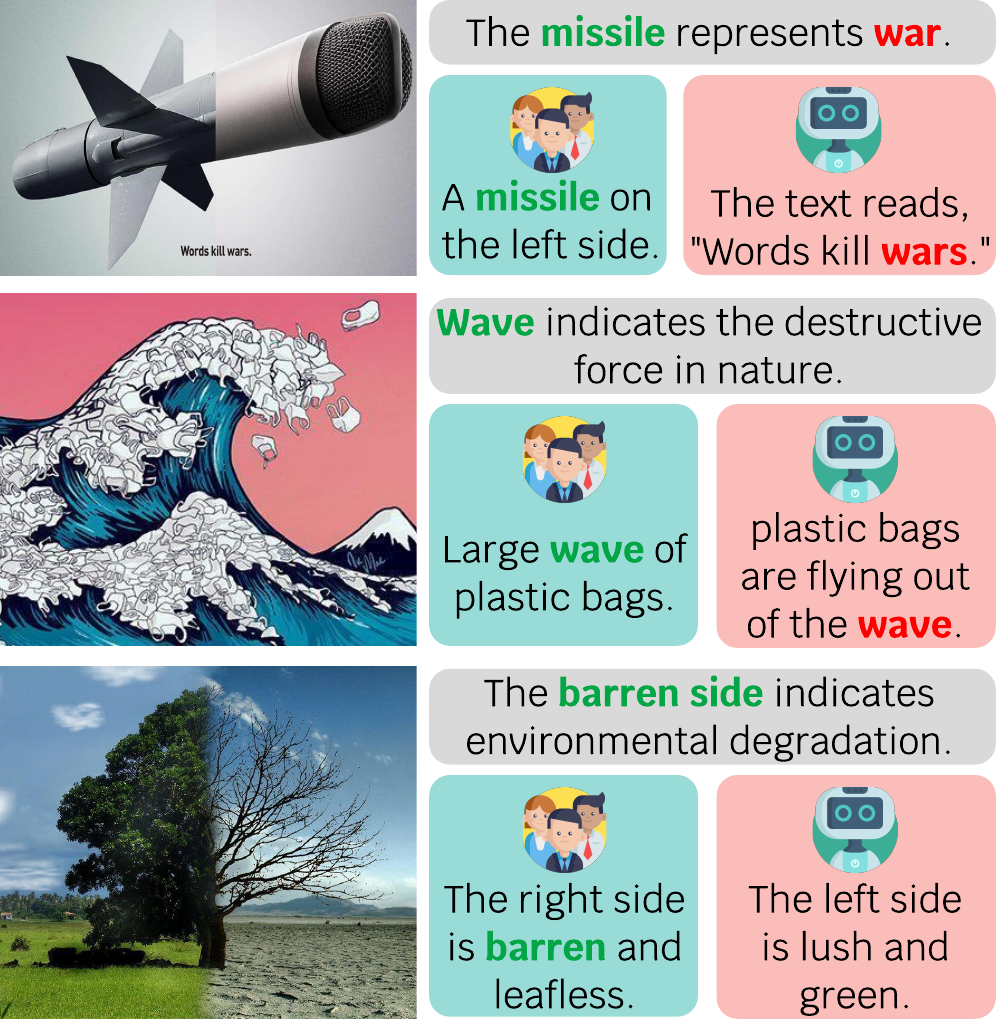}
    \caption{Failure cases of LLaVA-1.5 in \textit{Identification of Premises}. The model incorrectly reasons about relevant objects, relying instead on common words.}
    \label{fig:error}
\end{figure}

\begin{table}[!t]
    \centering
    \small
    \resizebox{0.48\textwidth}{!}{
    \begin{tabular}{l|ccccc}
    \toprule
    & Image & $\Delta$ VP & $\Delta$ CP & $\Delta$ Tree \\
    \midrule
    LLaVA-1.5 & 3.48 & $\uparrow$ \textbf{13.29} (5.42) & $\uparrow$ 8.57 (4.75) & $\uparrow$ 4.72 (4.34) \\
    LLaVA-Next & 15.04 & $\uparrow$ \textbf{11.28} (1.21) & $\uparrow$ 6.72 (2.78) & $\uparrow$ 4.14 (4.02) \\
    \bottomrule
    \end{tabular}
    }
    \caption{Mean of incremental improvements in BERTScore with each additional input across four different prompts in \textit{Deduction of Conclusion}. Standard deviations are shown in parentheses.}
    \label{tab:prompt_task3}
\end{table}

\subsection{Deduction of Conclusion}
\label{subsec:exp_con}

\textit{Deduction of Conclusion} evaluates the comprehensive ability to deduce the conclusion of an argument.
Given a subset of inputs among the image $I$, the visual premises $\textsc{VP}$, the commonsense premises $\textsc{CP}$, and the reasoning tree $T$, the objective is to generate the conclusion $\textsc{C}$ of an argument.


\bfsection{Metrics and Models}
As discussed earlier in~\cref{subsec:task_conclusion}, we use BERTScore as the primary metric. We supplement this with three additional static metrics (Bleu-4, ROUGE-L, CIDEr) in~\cref{sec:ax_full_results}.
The models tested in this task include all the multimodal LLMs used in the previous experiment and text-only LLMs (LLaMa-3-Instruct~\cite{llama3modelcard}, Mistral-Instruct~\cite{jiang2023mistral}, and Zephyr~\cite{tunstall2023zephyr}). All LLMs considered here are the $7\sim8$b sized variants.
The LLMs do not take the image as an input.

\bfsection{Results}
Table \ref{tab:task3_small} shows the results for this task. As expected from previous tasks, most models experience the highest gain from the additional information provided by the ground-truth set of visual premises.
This supports our hypothesis that selective attention to visual premises is a bottleneck in understanding visual arguments in current models. 
Also, both multimodal and text-only models benefited from commonsense premises and reasoning trees in most setups, indicating that models cannot yet perfectly understand visual arguments in a text-only format and benefit from explicit reasoning process information. 
We note that OFA struggled to follow the instruction format, leading to sub-zero scores. Although rare, BERTScore, based on cosine similarity, can yield negative values.
We also clarify that the multimodality of the \textit{deduction of conclusion} task resides in the visual premises, making it solvable by text-only models given them.

\subsection{Diagnostics}
\label{subsec:exp_ablation}

\bfsection{Prompt Robustness}
To ensure the robustness of our empirical results, we differentiated the prompts provided to the models. 
As shown in \cref{tab:prompt_task3}, the trend of gains remained stable across four different prompts, confirming the validity of our tests. For detailed prompts, refer to \cref{sec:ax_task3}, and for results in other tasks, see \cref{sec:ax_prompt_robustness}. 


\bfsection{Error Analysis}
\cref{fig:error} provides qualitative examples of failure cases. We present straightforward instances to clearly explain the errors. In these cases, the models fail to reason about the relevant object, which is the subject of the given intermediate conclusion, and instead rely on common words, leading to incorrect inference results.

\begin{figure}[!t]
    \includegraphics[width=0.48\textwidth]{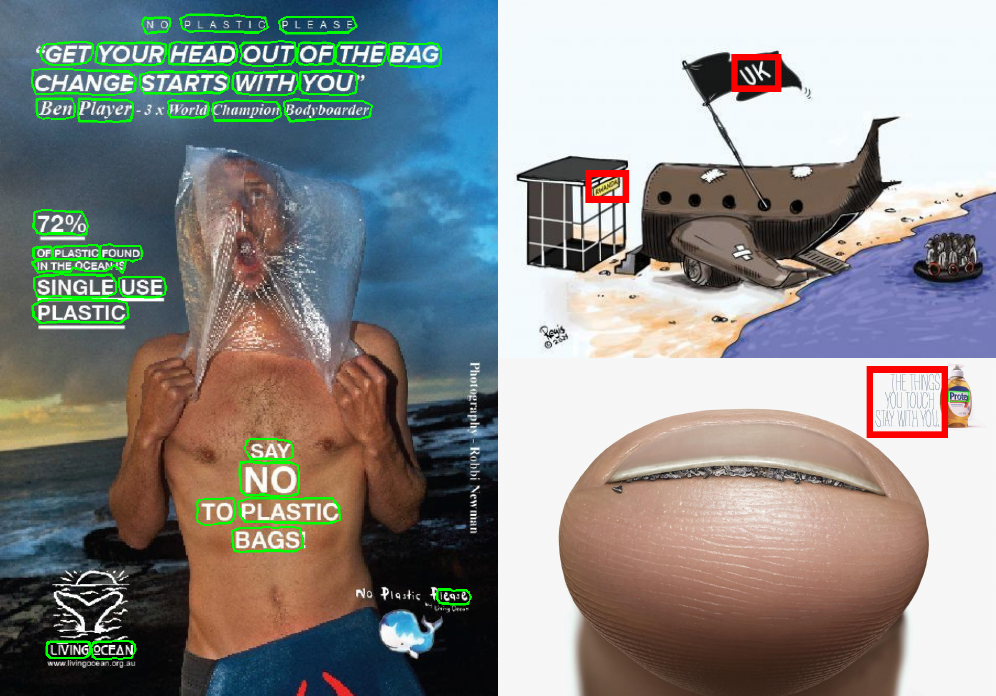}
    \caption{Left: OCR detection results. Right: Ground truth text instances missed by the model (highlighted in red). Most detection errors are attributed to Out-of-Domain cases such as calligraphy, handwritten text, or text that is too small for the model to detect, despite being distinguishable to the human eye.}
    \label{fig:ocr_fails}
\end{figure}

\bfsection{Reliance on OCR Capabilites}
To examine how demanding \modelname is on OCR capabilities, we employ a lightweight OCR detector~\cite{du2021pp} to detect bounding boxes of the visual premises \textit{without leveraging text annotations as input}. Even this simplified model achieves $82.77\%$ accuracy in image-wise evaluation, where an image is considered correctly detected only if all visual premises within it are identified. Typical failure cases are illustrated in~\cref{fig:ocr_fails}.

\section{Conclusion}
\label{sec:conclusion}

We introduce \modelname, a 
curated and annotated benchmark for visual argument understanding. Using our benchmark, we affirm a compelling hypothesis: selective vision is a critical bottleneck for visual reasoning in current machines. We aim for our benchmark to serve as a resource for advancing multimodal intelligence beyond passive captioning.
Future work includes:
\begin{enumerate}[leftmargin=*]
\setlength{\itemsep}{0pt} 
\setlength{\parskip}{0pt} 
\setlength{\topsep}{0pt} 
    \item Conditional Saliency Analysis: It is demonstrated that the saliency required for visual arguments differs from that needed for passive captioning. Can the varying saliency requirements across different tasks be analyzed?
    \item Extending Modalities: In speech recognition, non-conditional selective attention is known as the cocktail party effect. Would conditional selective attention be necessary in modalities other than vision as well?
\end{enumerate}

\section*{Limitations}
\label{sec:limitations}

\modelname, which is built on advertisements and cartoons from web sources, does not encompass all forms of visual arguments. Visual arguments also include various forms of media including mathematical diagrams~\cite{inglis2009persuasiveness} and videos, such as films~\cite{alcolea2009visual}. Consequently, the findings of this study do not represent all forms of visual arguments.

Additionally, the annotations for \modelname are created by two NLP researchers with similar cultural backgrounds. Although a different group of human evaluators validated these annotations, future research should consider individual variances in the interpretation of visual arguments and the reasoning processes identified by reasoning trees.

Finally, we excluded images containing written text in non-English languages when curating \modelname, as the annotators were not familiar with other languages. This limitation may confine the cultural context covered by \modelname, thus representing only a partial depiction of visual arguments. Since the logical relations forming a visual argument can depend on culture-specific elements, this skewed distribution of images can lead to a biased understanding of visual arguments.

We encourage future research to extend this work by exploring a wider range of visual arguments and incorporating more diverse cultural and linguistic contexts.

\section*{Acknowledgment}
\label{sec:ack}

This work was partly supported by an IITP grant funded by the Korean Government (MSIT) (No. RS-2020-II201361, Artificial Intelligence Graduate School Program (Yonsei University) and RS-2024-00353131) and the National Research Foundation of Korea (NRF) grant funded by the Korea government (MSIT) (No. RS-2024-00354218).

\bibliography{custom}

\clearpage

\clearpage

\appendix

\section{Data Annotation Details}
\label{sec:ax_data}

\bfsection{Human Resources}
To ensure a comprehensive understanding of the intricate requirements of our setup and maintain consistency across annotations, two of this paper's authors conducted the entire annotation process. Three volunteers from the NLP research community did the human evaluation.

\begin{figure*}
    \centering
    \includegraphics[width=\textwidth, keepaspectratio]{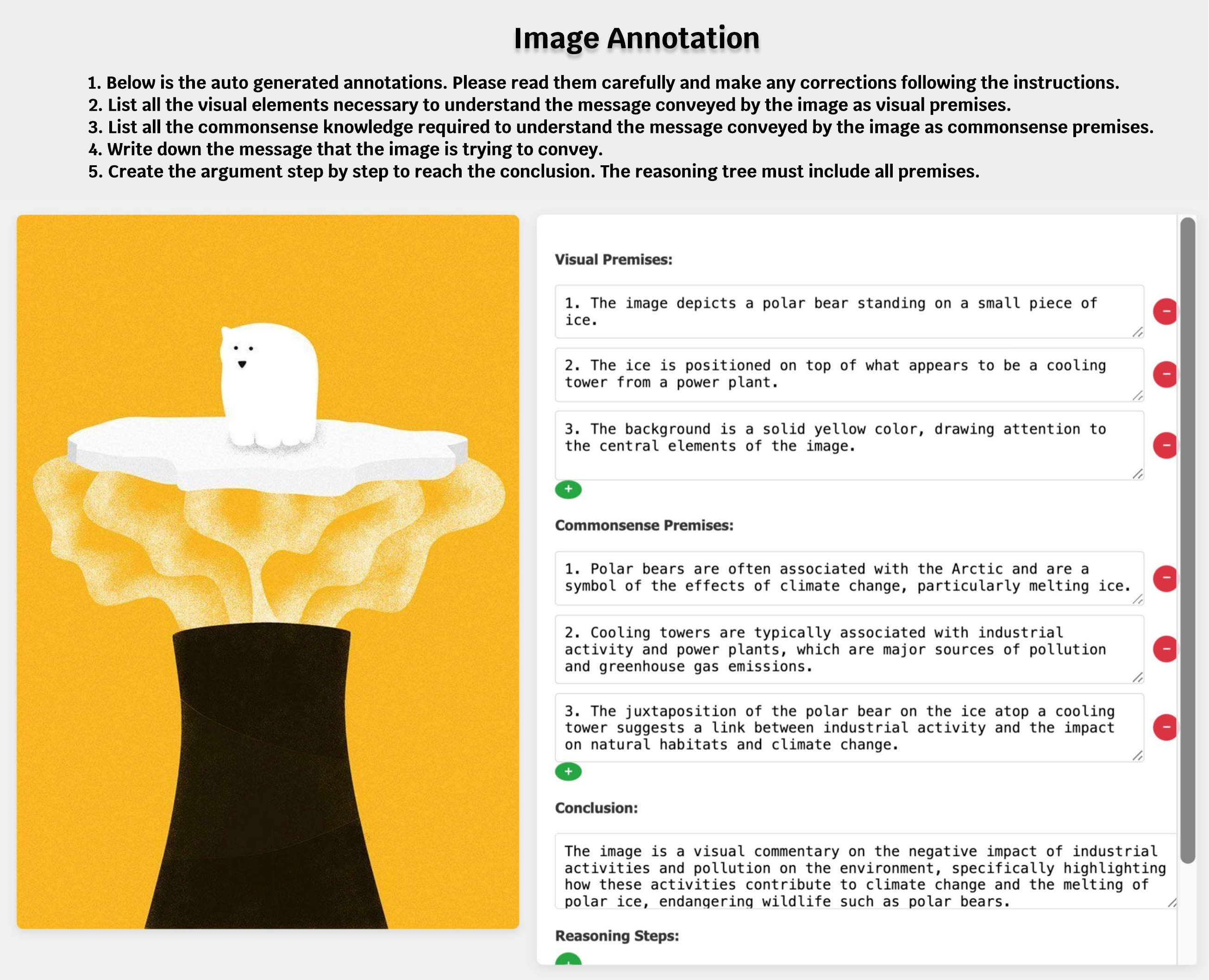}
    \caption{Human annotation interface for collecting textual annotations.}
    \label{fig:annotation1}
\end{figure*}

\begin{figure*}
    \centering
    \includegraphics[width=\textwidth, keepaspectratio]{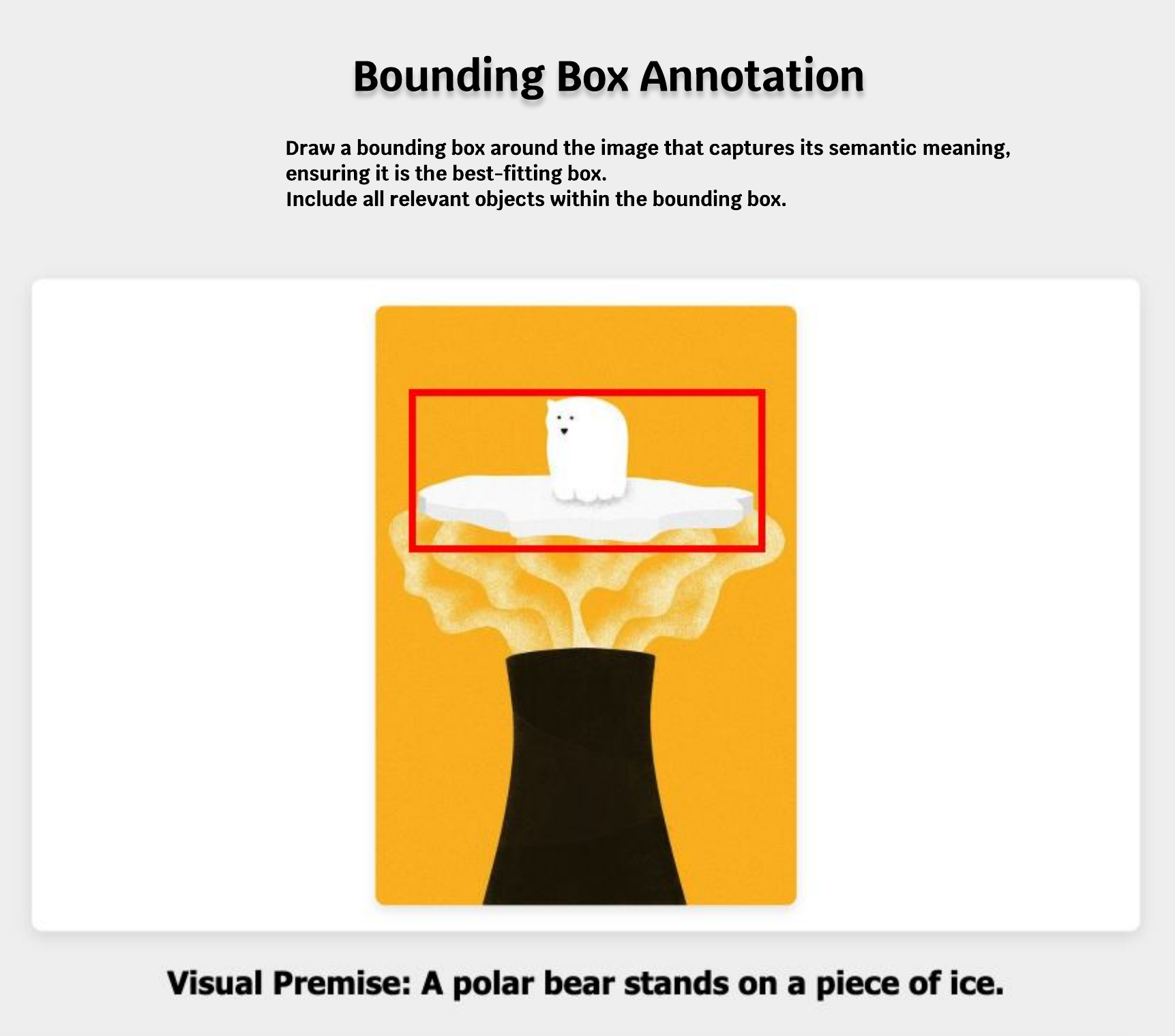}
    \caption{Human annotation interface for collecting bounding boxes of visual premises.}
    \label{fig:annotation2}
\end{figure*}

\begin{figure*}
    \centering
    \includegraphics[width=0.8\textwidth, keepaspectratio]{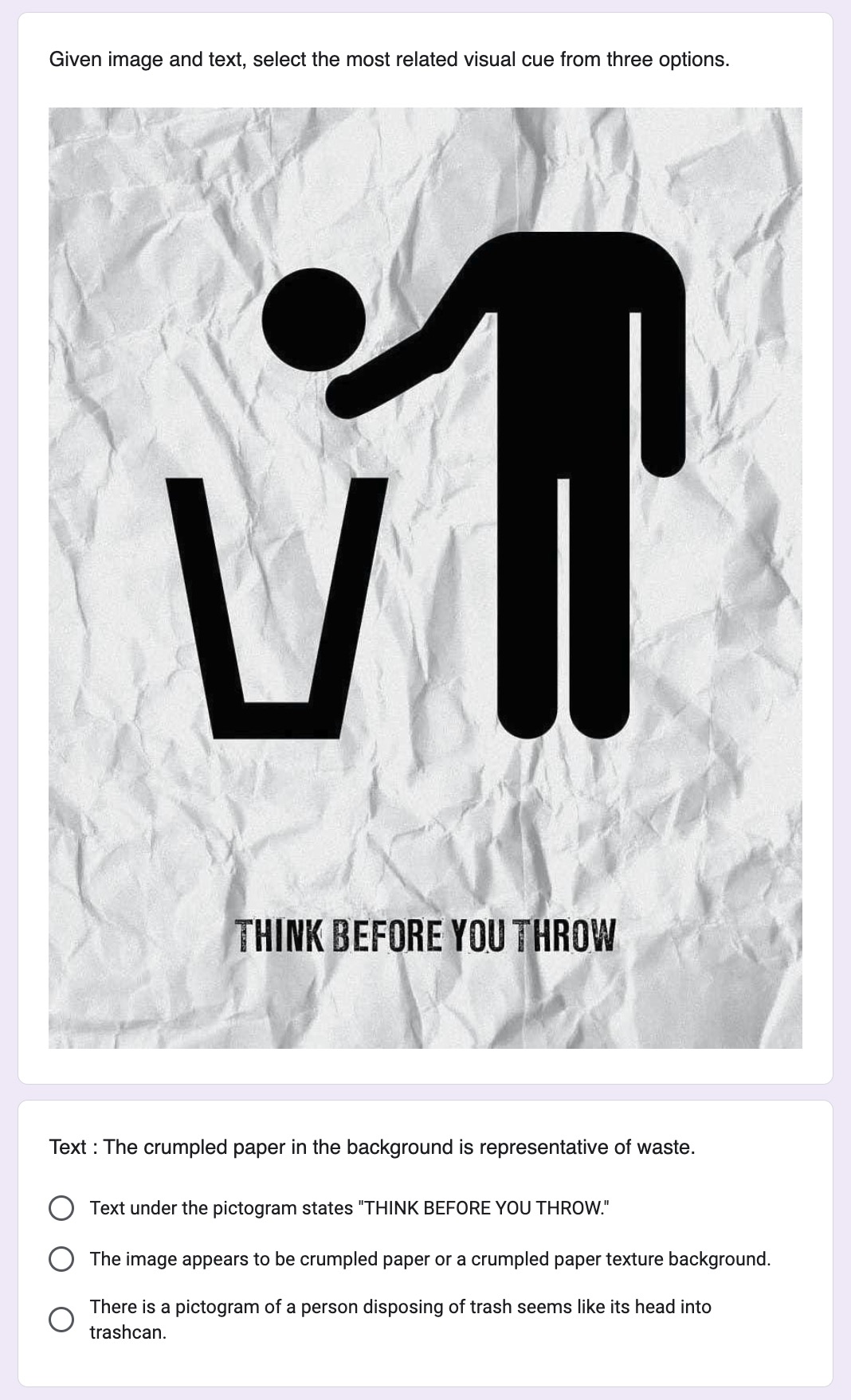}
    \caption{Human evaluation interface for \textit{Identification of Premises}.}
    \label{fig:task2_human_eval}
\end{figure*}

\bfsection{Annotation Interface}
We used a custom-built interface for efficient and convenient image annotation. The interface is depicted in~\cref{fig:annotation1} and ~\cref{fig:annotation2}. Additionally, we provide a snapshot of the human evaluation interface for \textit{Identification of Premises} in~\cref{fig:task2_human_eval}. We will open-source this interface along with the dataset.

\bfsection{Inter-Annotator Agreement}
The dataset annotation was conducted by two primary human annotators, with a third evaluator assigned to assess annotation quality and consistency. To measure inter-annotator agreement, 100 samples were randomly selected and re-annotated by a secondary annotator different from the original. Subsequently, the third evaluator independently assessed the equivalence of each annotated sample.

Given that the annotations comprise premise sets and natural language conclusions, rather than numerical scores, traditional metrics such as Cohen’s kappa are not applicable. Instead, we measured inter-annotator agreement using two distinct criteria: equivalence of visual premise sets and equivalence of natural language conclusions. The \textit{Jaccard similarity index} was employed to quantify the overlap between visual premise sets, while a binary equivalence test was used to evaluate alignment in the conclusions.

The results, as presented in~\cref{tab:ax_annotator_agreement}, demonstrate a high degree of inter-annotator agreement. Qualitative analysis revealed that observed discrepancies primarily stemmed not from substantive semantic variations but from differences in how annotators segmented a single concept into one or more visual premises.

\section{Analyzing Topic Diversity}
\label{sec:ax_topic}

Initially, we considered using the Latent Dirichlet Allocation (LDA)~\cite{blei2003latent} method for data visualization, following previous literature~\cite{hessel2022abduction}. However, we found that LDA based on Bag-of-Words representations could not generate meaningful clusters or labels for conclusion topics. As a solution, we developed an adaptive semantic classification technique using multimodal large language models:

\bfsection{Defining Class Labels} We utilize GPT-4-O. We first sample 400 sentences each for $\textsc{VP}$ and $\textsc{C}$, and then feed them to GPT with the following instructions:
For $\textsc{VP}$: \textit{"Give me well-balanced 10 object type classes for these texts (e.g., eating \& dining, environments \& landscapes, attire). Just classes."}
For $\textsc{C}$: \textit{"Give me well-balanced 10 classes for these texts. Just classes."}
After receiving the 10 classes from the GPT, we manually refine these classes into 8 classes for both $\textsc{VP}$ and $\textsc{C}$.

\bfsection{Labelling Data} We use a pretrained language model to classify visual premises ($\textsc{VP}$) and conclusions ($\textsc{C}$) in a zero-shot manner. We provide the following input to the LLaMA-3\footnote{\url{meta-llama/Meta-Llama-3-8B-Instruct}} LLM:
\begin{lstlisting}
Classes: {} 
Your task is to classify a sentence into 
the given classes. 
Give me just the class. 
Sentence: {} 
\end{lstlisting}

\bfsection{Visualization} We use the Plotly~\cite{plotly} library.
\section{Experiment Details}
\label{sec:ax_exp}

\begin{table}[t]
    \centering
    \small
    \begin{tabular}{l|cc}
    \toprule
    & Visual Premise &	Conclusion \\
    \midrule
Human-Human&	0.78&	0.96 \\
Machine-Machine&	0.51&	0.88 \\
    \bottomrule
    \end{tabular}
    \caption{Inter-annotator agreement results based on human evaluation, quantified using the Jaccard Similarity Index for set-level comparisons.}
    \label{tab:ax_annotator_agreement}
\end{table}

\begin{table}[t]
\centering
\small
\resizebox{1.0\linewidth}{!}{%
\begin{tabular}{l|cccc}
\toprule
model & dtype & \#parameter & version \\
\midrule
CLIP & - & 623M & RN50x64\\
CLIP & - & 427M & ViT-L/14 \\
CLIP & - & 427M & ViT-L/14@336px \\
SigLIP & - & 652M & large-patch16-384 \\
AlphaCLIP & - & 428M & clip\_l14\_336\_grit\_20m\_4xe \\
UNINEXT-H & - & 775M & image\_joint\_vit\_huge\_32g \\
LISA & - & 7B & xinlai/LISA-7B-v1\\
MM-G-DINO & - & 343M & grounding\_dino\_swin-l\_pretrain\_all\\
LLaVA-1.5 & FP16 & 7B & llava-1.5-7b-hf \\
LLaVA-NeXT & FP16 & 7B & mistral-v0.2 \\
Idefics2 & FP16 & 8B & chatty \\
OFA & - & 470M & vqa-pretrain-large \\
QwenVLChat & BF16 & 9B & Qwen-VL-Chat \\
CogVLM & BF16 & 17B & cogvlm-chat-hf \\
InstructBLIP & FP16 & 7B & instructblip-vicuna-7b  \\
Unified-IO-2 & - & 3B & uio2-xl \\
\bottomrule
\end{tabular}}
\caption{Details on the models used in our experiments.}
\label{tab:model_settings}
\end{table}

\subsection{Localization of Premises}

For closed-set grounding, we utilized CLIP, SigLIP, AlphaCLIP, and OFA. We measured the alignment between regions and descriptions of visual premises using image-to-text cosine similarity scores. The input regions were provided as cropped images. A model output was considered correct (True) if the similarity between the ground-truth region and the given description was the highest among all candidates; otherwise, it was marked incorrect (False).

For open-set grounding, we employed object grounding models such as MM-GDINO, UNINEXT-H, LISA, OFA, and Unified-IO-2 to directly generate bounding box coordinates. We applied a threshold of 0.35 to the outputs, merging the selected regions into the tightest rectangle union.
For LISA, we converted the output segmentation mask into bounding boxes. We then calculated the Intersection over Union (IoU) score for each bounding box. To compute the accuracy metric, we used a threshold of 0.5 for binary classification over the IoU. We calculated the local mean, which is the mean per visual premise in an image, and the mean per image.

\subsection{Identification of Premises}
We utilized OFA, Qwen-VL-Chat, CogVLM, Idefics2, InstructBLIP, Unified-IO-2, LLaVA-1.5, LLaVA-Next, and GPT-4-O for our experiments. We created multiple-choice questions with three possible answers: one correct answer and two incorrect answers. Five conditions were set for sampling the negatives for incorrect answers:
\begin{itemize}[leftmargin=*,topsep=0pt]
  \setlength{\itemsep}{0pt} 
  \setlength{\parskip}{0pt} 
  \setlength{\topsep}{0pt} 
    \item \textit{Random Sampling:} This global sampler selects samples uniformly without duplication.
    \item \textit{Visual Sampling:} This global sampler chooses the top 2 premise descriptions most similar to the image, using CLIP to score the cosine similarity between the image and text. We set the CLIP similarity threshold to 0.24 to ensure negative premises do not accurately describe the image.
    \item \textit{Textual Sampling:} This sampler selects the top 2 premise descriptions most similar to the ground truth premise, using ColBERT to score the cosine similarity between texts. We set the ColBERT similarity threshold to 25 to prevent negative premises from accurately describing the image.
    \item \textit{Mixed Sampling:} This approach combines visual and textual sampling, visually selecting from the top 10 textual retrieval results.
\end{itemize}

To ensure a fair comparison across various negative sampling methods, we use only intermediate conclusions that have three or more related visual premises. This results in 1,775 visual premises for the advertisement category and 1,774 for the cartoon category, totaling 3,549 visual premises, which is 62.34\% of the overall visual premises.


\bfsection{Human Evaluation}
We randomly selected 100 images from each data category and had human annotators perform the same tests as the machines across all negative set setups. The results demonstrated that humans achieved nearly perfect accuracy in this task, as shown in~\cref{tab:full_result_identification_of_premises}.


\subsection{Deduction of Conclusion}
We conducted experiments on both Multi-Modal Large Language Models (MLLM) and Large Language Models (LLMs). The MLLMs used in our experiments include LLaVA-1.5, LLaVA-NeXT, Idefics2, OFA, InstructBLIP, Qwen-VL-Chat, CogVLM, and Unified-IO-2. The LLMs include LLaMA-3, Mistral, and Zephyr.

\bfsection{Prompting}
Before conducting the experiments, we established a set of instructions to be applied to all models to elicit appropriate responses. During this process, we encountered several issues with prompt engineering, such as model refusal to address controversial or unsafe questions, the inclusion of unnecessary tokens, multiple sentences, and the positioning of image tokens.
Ultimately, we decided on the following prompt: \textit{"<image> <information> Your task is to answer what the image wants to convey. You should respond in only one sentence without any unnecessary prefixes. ANSWER:"}

\subsection{Resource \& Hyperparameters}

\bfsection{Computation}
We utilized RTX-4090 and A6000 GPUs for our experiments. All models, except for CogVLM, were implemented using RTX-4090 GPUs. Due to the size of its model weights, CogVLM was implemented on an A6000 GPU. Each model required up to 8 RTX-4090 GPU-hours per task. In total, conducting all tasks demanded 200 RTX-4090 GPU-hours.

\bfsection{Hyperparameters}
Our experiments are deterministic, given the pretrained model weights, the greedy decoding scheme, and the instruction prompts. We explore prompt diversification in~\cref{subsec:exp_ablation} and~\cref{sec:ax_prompt_robustness}.

\subsection{Model Details}

We specify all exact model identifiers and sizes in~\cref{tab:model_settings}.

\label{sec:ax_exp_model}

\section{Comparison of Metrics for \textit{Deduction of Conclusion}}
\label{sec:ax_score_comparison}

Here, we describe details for human evaluation of goodness per each metric illustrated in~\cref{tab:score_comparison}.

\bfsection{Human Evaluation}
We sampled 200 target images and collected responses from three models: LLaVA-Next, Qwen-VL-Chat, and GPT-4o. Human annotators then determined whether each model's conclusion was semantically similar to the reference conclusion.

\bfsection{Metrics}
To evaluate accuracy, precision, recall, and F1-score, we first converted each metric into binary decisions using derived thresholds. We established these thresholds by training a logistic regression model on 100 pairs of metric scores and human decisions. Subsequently, we inferred binary decision labels on the remaining 100 pairs. The results are presented in~\cref{tab:score_human_eval}. Additionally, the correlation between the metrics and human decisions is reported using Pearson's coefficient~\cite{cohen2009pearson}.
\begin{table*}[ht]
\centering
\small
\begin{tabular}{l|c|ccc|c}
\toprule
& Accuracy & Precision & Recall & F1-score & Pearson Corr. ($\rho$) \\
\midrule
BLEU-4 &0.67 & 0.44 & 0.67 & 0.53 & 0.18 \\
ROUGE &0.75 & 0.76 & 0.75 & 0.72 & 0.35\\
CIDEr &0.72 & 0.70 & 0.72 & 0.70 & 0.26 \\
GPTEval &0.75 & 0.83 & 0.75 & 0.76 & 0.53 \\
\textbf{BERTScore} &\textbf{0.94} & \textbf{0.94} & \textbf{0.93} & \textbf{0.93} & \textbf{0.59} \\
\bottomrule
\end{tabular}
\caption{Comparison of metrics with human decision on \textit{Deduction of Conclusion}}
\label{tab:score_human_eval}
\end{table*}


\section{Prompt Robustness in \textit{Identification of Premises}}
\label{sec:ax_prompt_robustness}
\begin{table*}[t]
    \small
    \centering
    \begin{tabular}{l|c|cccc|c}
    \toprule
    &  & \multicolumn{4}{c|}{Global} & \multicolumn{1}{c}{Local} \\
    \midrule
    & Prompt & Random & Visual & Textual & Mixed & \multicolumn{1}{c}{Semantic} \\
    \midrule
    LLaVA-NeXT & \multirow{2}{*}{Original} &97.10 & 96.14 & 80.53 & 84.70 & 77.51 \\
    InstructBLIP &  &90.65 & 84.53 & 71.54 & 74.75 & 58.21 \\
    \midrule
    LLaVA-NeXT & \multirow{2}{*}{Paraphrase 1} & 97.24 & 96.59 & 80.98 & 85.63 & 77.60 \\ 
    InstructBLIP &   & 90.93 & 84.73 & 72.78 & 74.98 & 59.68 \\ 

    \midrule
    LLaVA-NeXT & \multirow{2}{*}{Paraphrase 2} & 97.60 & 96.25 & 81.46 & 86.00 & 76.67 \\ 
    InstructBLIP &   & 93.32 & 89.83 & 79.01 & 81.71 & 64.50 \\     
    \bottomrule
    \end{tabular}
    \captionof{table}{Assessment of prompt robustness with different paraphrases in \textit{Identification of Premises}. Accuracy is measured as a percentage.}
    \label{tab:task2_prompt_robust}
\end{table*}

\begin{table}[h]
    \centering
    \small
    \resizebox{1.0\linewidth}{!}{%
    \begin{tabular}{l|cccccc}
    \toprule
    & \multicolumn{3}{c}{IoU} & \multicolumn{3}{c}{Acc. ($\%$)} \\
    \midrule
    & Ads & Cartoon & All & Ads & Cartoon & All \\
    \midrule
    UNINEXT-H & 34.50 & 44.33  &38.75 & 31.67 & 40.71&  35.58  \\
    LISA & 40.05 & 49.17 & 44.25 & 40.52 & 50.01 & 44.62 \\
    Unified-IO-2 & 45.81 & 52.29 &48.61 & 44.66 & 50.43 & 47.15  \\
    OFA & 49.10 & 51.49 &50.14 & 49.06 & 49.22 & 49.13  \\
    MM-G-Dino & 52.70 & 58.06 &55.02 & 52.39 & 58.37 & 54.98 \\
    \bottomrule
    \end{tabular}%
    }
    \caption{\textit{Open-set} grounding results in \textit{localization of premises}.}
    \label{tab:task1_open_full}
\end{table}

Extending the robustness study in~\cref{tab:prompt_task3}, 
we conducted a similar prompt diversification experiment for the task of \textit{Identification of Premises}. By paraphrasing the original prompt as described in~\cref{sec:ax_task2}, we performed the same evaluation. The results, presented in~\cref{tab:task2_prompt_robust}, demonstrate that our experimental outcomes remain stable for \textit{Identification of Premises} across different prompt paraphrases.

\section{Full Results}
\label{sec:ax_full_results}

This section presents the comprehensive versions of the results summarized in the main paper. \cref{tab:task1_open_full} displays the \textit{open-set} grounding results for \textit{Localization of Premises}, while \cref{tab:full_result_identification_of_premises} provides the results for \textit{Identification of Premises}. The results for the task of \textit{Deduction of Conclusion} are detailed by category: advertisements are shown in \cref{tab:ads_conclusion_deduction}, cartoons in \cref{tab:cartoon_conclusion_deduction}, and the average across both categories in \cref{tab:total_conclusion_deduction}.

\section{Qualitative Samples on \textit{Open-Set} Grounding}
\label{sec:ax_open_grounding_samples}

\begin{figure*}
    \includegraphics[width=\textwidth, keepaspectratio]{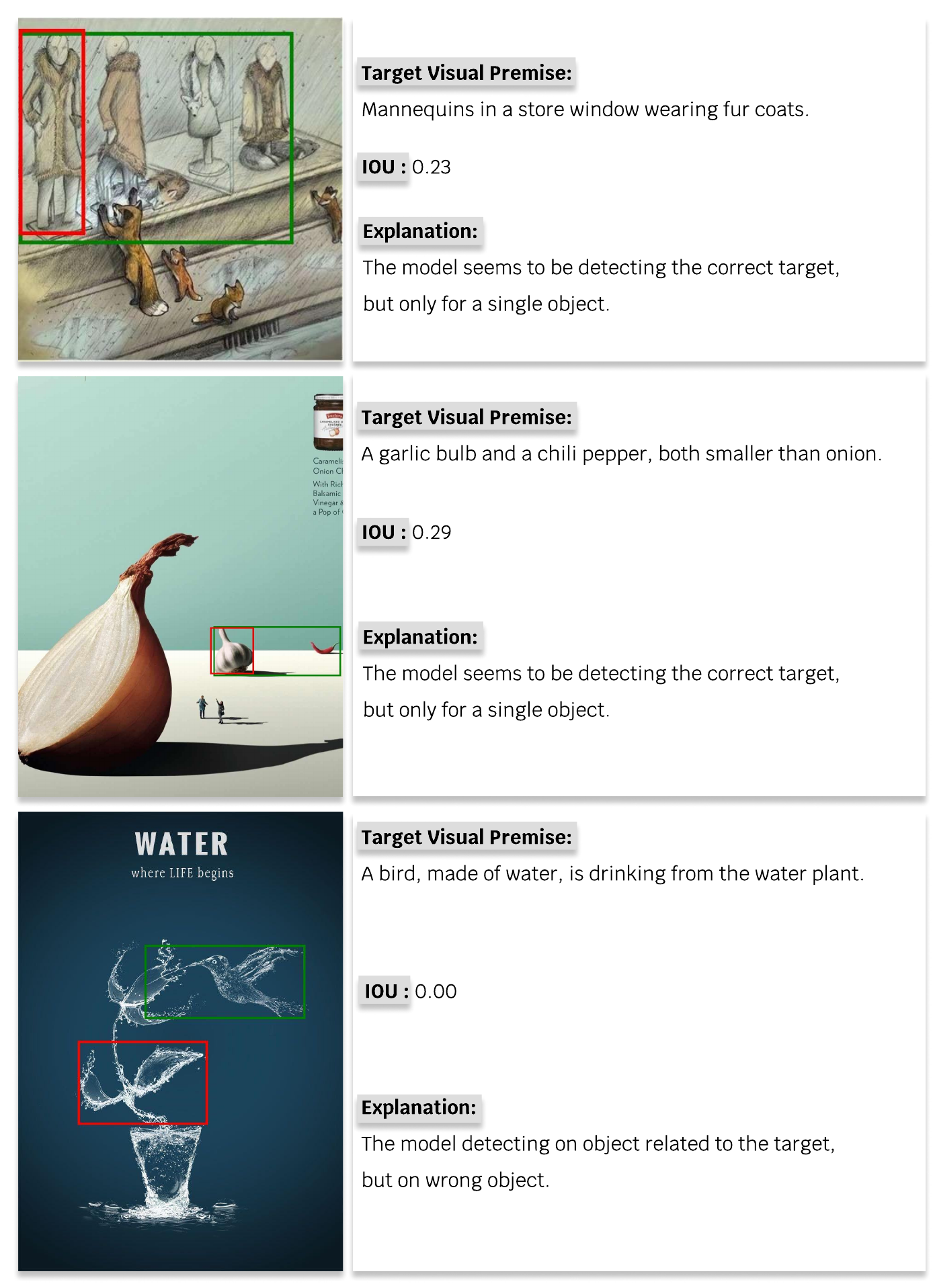}
    \caption{Qualitative samples of \textit{open-set} grounding results.}
    \label{fig:qualitative_sample_openset_grounding}
\end{figure*}

To identify the cause of low performance in the \textit{open-set} evaluation of the \textit{Localization of Premises} task, we examine qualitative samples shown in \cref{fig:qualitative_sample_openset_grounding}. Traditional object detection models are typically trained on single object labels, whereas our semantic region labels may encompass multiple objects with similar meanings. Consequently, although the models may detect the correct target, the intersection over union (IoU) scores are lower, resulting in reduced accuracy.

\section{Qualitative Samples on \textit{Deduction of Conclusion}}
\label{sec:ax_deduction_samples}

\begin{figure*}
    \includegraphics[width=\textwidth, keepaspectratio]{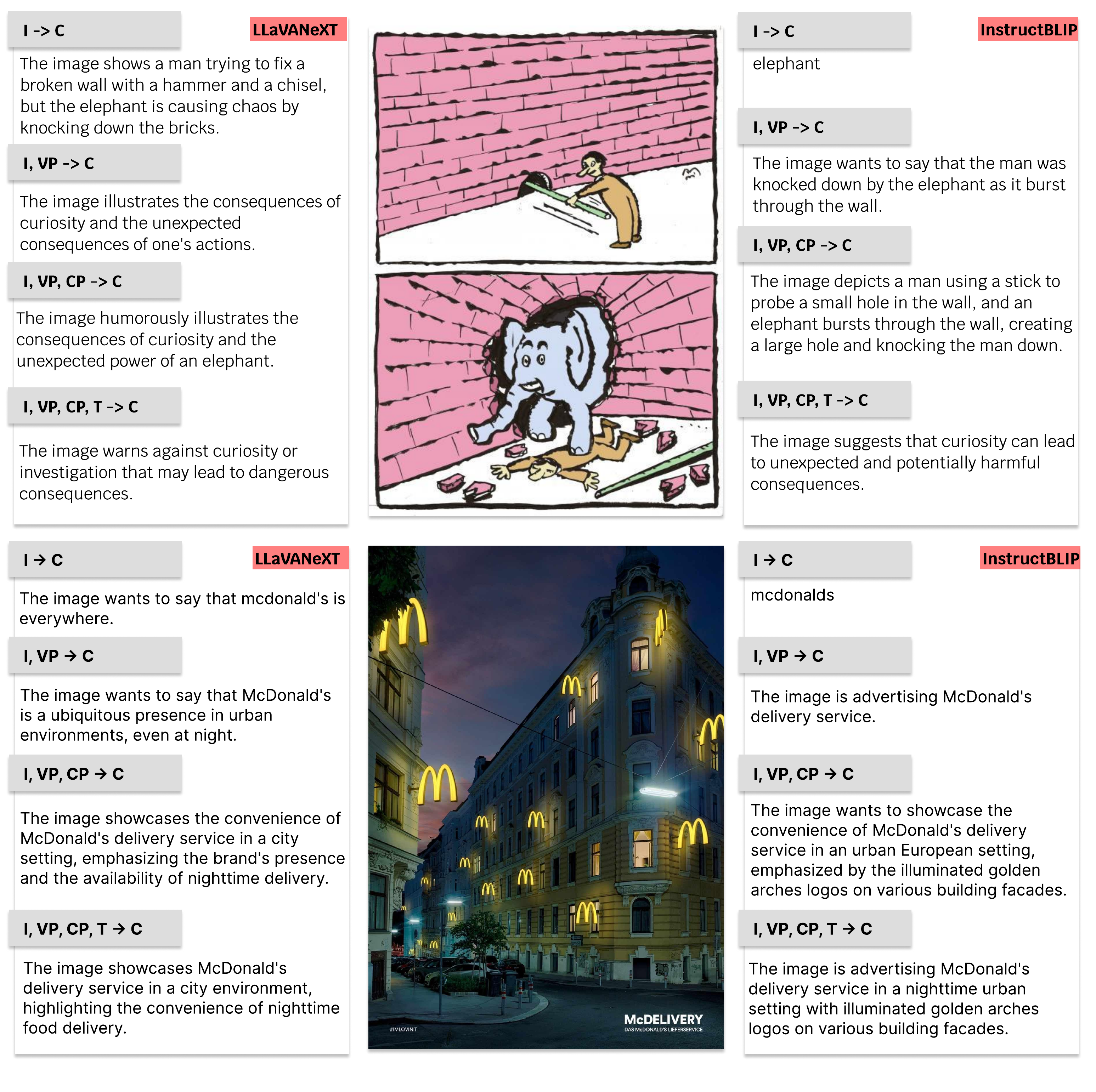}
    \caption{Qualitative samples of \textit{Deduction of Conclusion}, with inference results from LLaVA-NeXT and InstructBLIP models.}
    \label{fig:ax_deduction_of_conclusion_1}
\end{figure*}

\begin{figure*}
    \includegraphics[width=\textwidth, keepaspectratio]{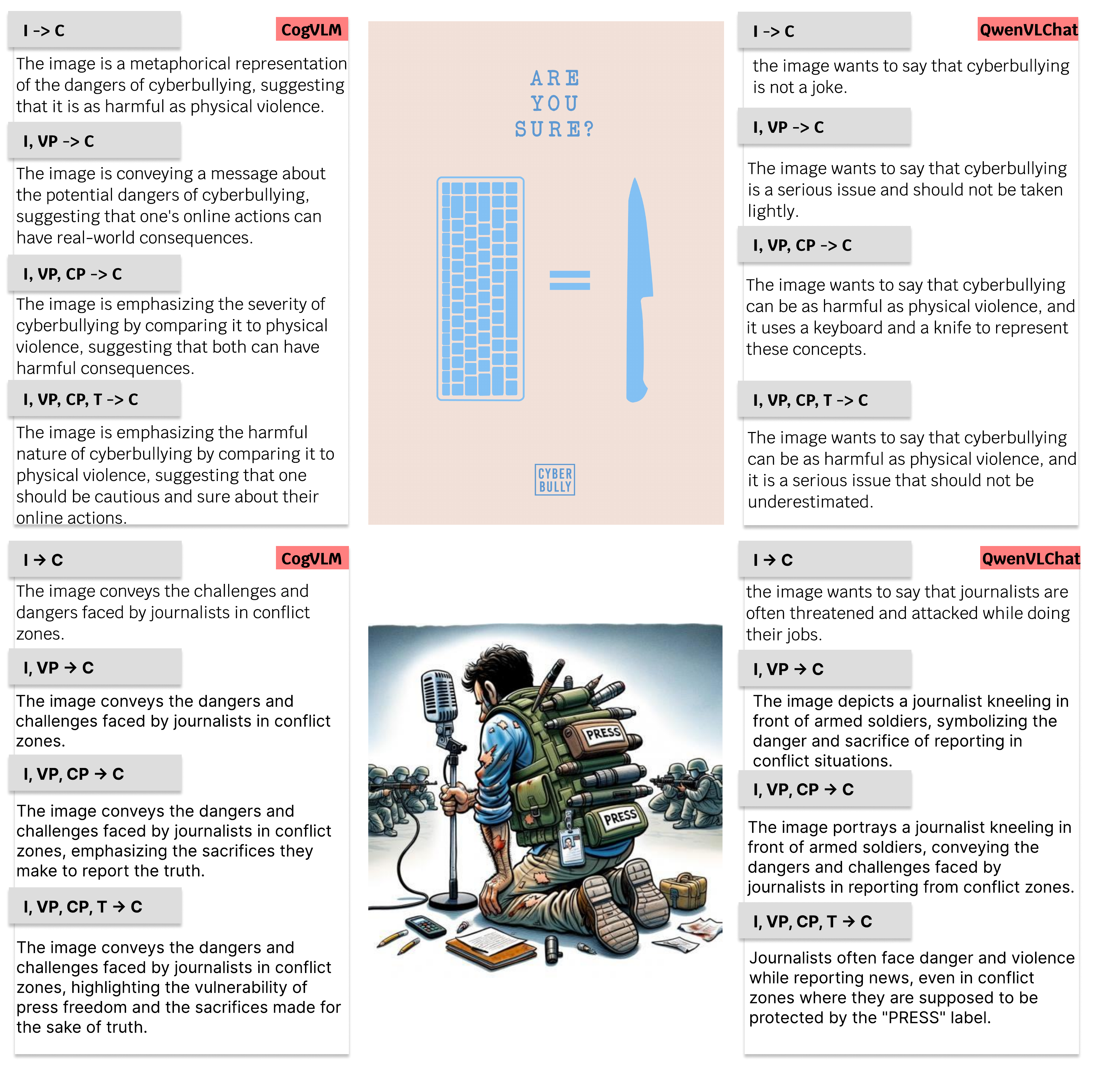}
    \caption{Qualitative samples of \textit{Deduction of Conclusion}, with inference results from CogVLM and Qwen-VL-Chat models.}
    \label{fig:ax_deduction_of_conclusion_2}
\end{figure*}

Inference results of different models with varying inputs are shown in \cref{fig:ax_deduction_of_conclusion_1} and \cref{fig:ax_deduction_of_conclusion_2}. The outputs of the models display discrepancies; for instance, CogVLM exhibits weak conditioning on additional inputs, producing similar outputs despite the incremental increase in information provided through different inputs.

\section{Credits}

We do not claim any rights to the images included in our dataset. Therefore, we provide only the URLs to the corresponding images instead of distributing the raw files. For usage outside of an academic context, please contact the copyright holders directly.

\bfsection{Figures}
All icons used in the figures are from \url{www.flaticon.com}.
\begin{itemize}[leftmargin=*,topsep=0pt]
  \setlength{\itemsep}{0pt} 
  \setlength{\parskip}{0pt} 
  \setlength{\topsep}{0pt} 
    \item Figure 1: \url{www.art-vibes.com/design/egle-plytnikaite-environmental-issues}
    \item Figure 2: \url{www.commarts.com/project/24399/mcdonald-s-refresh}
    \item Figure 3: \url{www.aisleone.net/2007/10/30/jeep/} | \url{www.nextml.github.io/caption-contest-data/dashboards/630.html} | \url{www.i.pinimg.com/originals/ac/32/16/ac321665c9e8f5feccc62eb3f6d09d37.jpg} | \url{www.ipnoze.com/publicite-sociale/} | \url{www.adsoftheworld.com/campaigns/scissors-1e569372-d5e7-488b-9e06-8bf46580801e}
    \item Figure 5: \url{www.adsoftheworld.com/campaigns/words-1c383606-d2b3-4aea-9f19-c0627b6fb4ff} | \url{www.behance.net/gallery/68747547/The-Great-Plastic-Wave} | \url{www.fanpop.com/clubs/global-warming-prevention/images/33088666/title/global-warming-photo}
\end{itemize}

\section{Prompts for Annotation}
\label{sec:ax_data_prompt}

\begin{itemize}

\item \textbf{Annotation for Visual Premises} \\
\hspace{0.0em}
\begin{lstlisting}
Your task is to identify visual premises from the image. These are visual cues that support or illustrate the conclusion, enhancing the overall understanding and clarity of the image.

Example
Visual Premises (VP):
1. The image depicts a maze with entry point and exit.
2. At the entry point of a maze labeled "Start," there is a cigarette.
3. The exit of the maze is labeled "Lung Cancer."
4. There's a text saying, "Or you can start here," with an arrow pointing to another text that reads, "Make the right choice. DON'T SMOKE."
\end{lstlisting}

\item \textbf{Annotation for Constructing Arguments} \\
\hspace{0.0em}
\begin{lstlisting}
Visual Premises (VP):
1. VP1
2. VP2
3. VP3

Given the visual premises of the image, your task is to generate the necessary commonsense premises and conclusion of the image. The conclusion should be one simple sentence. Then show the reasoning steps to reach the conclusion. The reasoning steps should include all visual premises and commonsense premises. You can refer to the following example.

Example
Visual Premises (VP):
1. The image depicts a maze with entry point and exit.
2. At the entry point of a maze labeled "Start," there is a cigarette.
3. The exit of the maze is labeled "Lung Cancer."
4. There's a text saying, "Or you can start here," with an arrow pointing to another text that reads, "Make the right choice. DON'T SMOKE."

Commonsense Premises (CP):
1. Mazes are often used to represent complex journeys or paths one must navigate.
2. Cigarettes are known to be harmful to health and a major cause of lung cancer.
3. The phrase "Make the right choice" implies that there is a decision to be made that can impact one's health.
4. Public health messages often use strong visuals to convey the importance of making healthy choices.

Conclusion (C):
The image is a public health message that illustrates the dangerous path from smoking to lung cancer while encouraging individuals to choose not to smoke for their health.

Reasoning Steps:
(VP1, CP1 -> IC1): The maze represents the difficult and potentially harmful journey.
(VP2, CP2 -> IC2): The presence of a cigarette at the maze's entry point indicates the start of this hazardous journey.
(VP3, CP2 -> IC3): Labeling the maze's exit as "Lung Cancer" directly links smoking to this deadly disease.
(VP4, CP3, CP4 -> IC4): The additional text offers an alternative choice to avoid smoking, emphasizing the importance of preventive health measures.
(IC1, IC2, IC3, IC4 -> C): The image is a public health message that warns about the risks of smoking and encourages making the right choice for one's health.

Answer
\end{lstlisting}
\end{itemize}

\section{Prompts for Evaluation}
\label{sec:ax_eval_prompt}

\begin{itemize}
\item \textbf{GPTEval} \\
\hspace{0.0em}
\begin{lstlisting}
Task Description: You will be given a ground truth sentence that describes an image and a model-generated sentence. Your task is to evaluate the semantic similarity between the model-generated sentence and the ground truth sentence. You don't need to give me any description. Just score should be answered.
Evaluation Criteria: T/F. False means the sentences are completely different. True means they mean exactly the same thing.

Ground Truth: {}
Generated: {}
\end{lstlisting}
\end{itemize}

\section{Prompts for \textit{Identification of Retrieval}}
\label{sec:ax_task2}
\begin{itemize}

\item \textbf{Original Prompt}\\
\hspace{0.0em}
\small
\begin{lstlisting}
<image>
The following are multiple choice questions (with answers) about image understanding.

When given an image, a conclusion, and several visual cue options, you need to identify the visual cue that best relates to the conclusion. To do this effectively, carefully analyze how each visual cue connects to the key elements of the conclusion. Select the visual cue that most directly supports or illustrates the conclusion, ensuring that it enhances the overall understanding and clarity of the message. Answer A), B), or C) with no additional explanation. Conclusion: {conclusion}
{vp_options}
ANSWER:
\end{lstlisting}

\item \textbf{Paraphrase 1}\\
\hspace{0.0em}
\small
\begin{lstlisting}
<image>
The following are multiple choice questions (with answers) about image understanding.

When given an image, a conclusion, and several visual cue options, identify the visual cue that best relates to the conclusion. Select the visual cue that most directly supports or illustrates the conclusion, ensuring that it enhances the overall understanding and clarity of the message. To do this effectively, carefully analyze how each visual cue connects to the key elements of the conclusion. Answer A), B), or C) with no additional explanation. Conclusion: {conclusion}
{vp_options}
ANSWER:
\end{lstlisting}

\item \textbf{Paraphrase 2}\\
\hspace{0.0em}
\small
\begin{lstlisting}
<image>
The following are multiple choice questions (with answers) about image understanding.

Given an image, what is the visual cue most related to the given conclusion? Answer A), B), or C) with no additional explanation. Conclusion: {conclusion}
{vp_options}
ANSWER:
\end{lstlisting}

\end{itemize}

\section{Prompts for \textit{Deduction of Conclusion}}
\label{sec:ax_task3}
\begin{itemize}

\item \textbf{Image -> C}\\
\hspace{0.0em}
\small
\begin{lstlisting}
<image>
Your task is to answer what the image wants to say. You should answer in only one sentence without an unnecessary prefix. ANSWER:
\end{lstlisting}

\item \textbf{Image, VP -> C}\\
\hspace{0.0em}
\small
\begin{lstlisting}
<image>
"Visual Premises (VP)" are the important features presented in the images.

Visual Premises (VP):
1. VP1
2. VP2
3. VP3

Your task is to answer what the image wants to say. You should answer in only one sentence without an unnecessary prefix. ANSWER:
\end{lstlisting}

\item \textbf{Image, VP, CP -> C}\\
\hspace{0.0em}
\small
\begin{lstlisting}
<image>
"Visual Premises (VP)" are the important features presented in the images. " Commonsense Premises (CP)" are not visually depicted in the image but are commonly understood by people.

Visual Premises (VP):
1. VP1
2. VP2
3. VP3

Commonsense Premises (CP):
1. CP1
2. CP2
3. CP3

Your task is to answer what the image wants to say. You should answer in only one sentence without an unnecessary prefix.
ANSWER:
\end{lstlisting}

\item \textbf{Image, VP, CP, Tree -> C}\\
\hspace{0.0em}
\small
\begin{lstlisting}
<image>
"Visual Premises (VP)" are the important features presented in the images. " Commonsense Premises (CP)" are not visually depicted in the image but are commonly understood by people. "Reasoning Steps" are the structure of explanation of how we came up to the "Intermediate Conclusion(IC) and "Conclusion".

Visual Premises (VP):
1. VP1
2. VP2
3. VP3

Commonsense Premises (CP):
1. CP1
2. CP2
3. CP3

Reasoning Step:
(VP1, CP1 -> IC1): IC1
(VP2, CP2 -> IC2): IC2
(VP3, CP3 -> IC3): IC3
(IC1, IC2, IC3 -> C):

Your task is to answer what the image wants to say. You should answer in only one sentence without an unnecessary prefix. ANSWER:
\end{lstlisting}

\item \textbf{Prompt Style 1}\\
\hspace{0.0em}
\small
\begin{lstlisting}
<image>
"Visual Premises (VP)" are the important features presented in the images. " Commonsense Premises (CP)" are not visually depicted in the image but are commonly understood by people. "Reasoning Steps" are the structure of explanation of how we came up to the "Intermediate Conclusion(IC) and "Conclusion".

Visual Premises (VP):
...

Commonsense Premises (CP):
...

Reasoning Step:
...

Answer in one sentence what the image wants to convey. ANSWER:
\end{lstlisting}

\item \textbf{Prompt Style 2}\\
\hspace{0.0em}
\small
\begin{lstlisting}
<image>
"Visual Premises (VP)" are the key visual elements in the image. "Commonsense Premises (CP)" are elements based on general common sense. "Reasoning Steps" are the process of reaching the "Intermediate Conclusion (IC)" and "Conclusion".

Visual Premises (VP):
...

Commonsense Premises (CP):
...

Reasoning Step:
...

Write the message of the image in one sentence. You should answer in only one sentence without an unnecessary prefix. RESPONSE:
\end{lstlisting}

\item \textbf{Prompt Style 3}\\
\hspace{0.0em}
\small
\begin{lstlisting}
<image>
"Visual Premises (VP)" represent the important features observed in the image. "Commonsense Premises (CP)" are things not visually depicted but generally understood. "Reasoning Steps" are the explanation process leading to the "Intermediate Conclusion (IC)" and "Conclusion".

Visual Premises (VP):
...

Commonsense Premises (CP):
...

Reasoning Step:
...

Write the main message of the image in one sentence. RESPONSE:
\end{lstlisting}

\item \textbf{Prompt Style 4}\\
\hspace{0.0em}
\small
\begin{lstlisting}
<image>
"Visual Premises (VP)" are the key features observed in the image. "Commonsense Premises (CP)" are not visually depicted but can be understood through general knowledge. "Reasoning Steps" are the logical explanation process leading to the "Intermediate Conclusion (IC)" and "Conclusion".

Visual Premises (VP):
...

Commonsense Premises (CP):
...

Reasoning Step:
...

Write the meaning the image wants to convey in one sentence. RESPONSE:
\end{lstlisting}

\end{itemize}

\begin{table*}[!h]\centering
    \centering
    \resizebox{1.0\linewidth}{!}{%
    \begin{tabular}{l|ccc|ccc|ccc|ccc|ccc}
    \toprule
    & \multicolumn{12}{c|}{Global} & \multicolumn{3}{c}{Local} \\
    \midrule
    & \multicolumn{3}{c}{Random} & \multicolumn{3}{c}{Visual} & \multicolumn{3}{c}{Textual} & \multicolumn{3}{c|}{Mixed} & \multicolumn{3}{c}{Semantic} \\
    \midrule
    & Ads & Cartoon & All & Ads & Cartoon & All & Ads & Cartoon & All & Ads & Cartoon & All & Ads & Cartoon & All \\
    \midrule
    Random & 33.33 & 33.33 & 33.33 & 33.33 & 33.33 & 33.33 & 33.33 & 33.33 & 33.33 & 33.33 & 33.33 & 33.33 & 33.33 & 33.33 & 33.33  \\
    Human & 100.00 & 100.00 & 100.00 & 100.00 & 98.00 & 99.00 & 96.00 & 92.00 & 94.00 & 100.00 & 100.00 & 100.00 & 98.00 & 98.00 & 98.00 \\
    \midrule
    OFA & 0.00 & 0.00 & 0.00 & 0.00 & 0.00 & 0.00 & 0.00 & 0.00 & 0.00 & 0.00 & 0.00 & 0.00 & 0.00 & 0.00 & 0.00  \\
    Qwen-VL-Chat & 88.90 &  83.21 &  86.05 &  88.67 &  82.87 &  85.77 &  73.73 &  67.61 &  70.67 &  77.00 &  74.14 &  49.73 &  53.21 &  46.25 &  75.57 \\
    CogVLM & 97.58 &  97.35 &  97.46 &  96.45 &  96.34 &  96.39 &  88.78 &  87.21 &  88.00 &  91.66 &  92.79 &  92.22 &  69.28 &  61.35 &  65.31 \\
    Idefics2 & 98.59 &  98.76 &  98.68 &  97.91 &  97.75 &  97.83 &  93.18 &  90.42 &  91.80 &  95.15 &  94.99 &  95.07 &  77.40 &  72.62 &  75.01 \\
    InstructBLIP & 82.41 &  85.13 &  83.77 &  78.07 &  80.39 &  79.23 &  68.55 &  65.35 &  66.95 &  71.87 &  70.87 &  71.37 &  66.91 &  56.90 &  61.90 \\
    Unified-IO-2 & 98.31 & 98.54 & 98.42 & 97.29 & 96.68 & 96.99 & 88.78 & 84.96 & 86.87 & 92.28 & 93.35 & 92.81 & 34.67 & 34.82 & 34.74 \\
    LLaVA-1.5 & 98.82 &  98.48 &  98.65 &  98.08 &  97.75 &  97.91 &  84.44 &  83.04 &  83.74 &  89.23 &  90.48 &  89.86 &  73.34 &  61.52 &  67.43  \\
    LLaVA-NeXT & 97.35 &  97.97 &  97.66 &  96.05 &  96.34 &  96.20 &  81.17 &  80.62 &  80.90 &  84.33 &  87.38 &  85.86 &  82.69 &  74.37 &  78.53  \\ 
    GPT-4-O &  - & - & - & - & - & - & - & - &- &-&-& -&75.22 &82.56&79.50 \\
    \bottomrule
    \end{tabular}%
    }
    \caption{Results on \textit{Identification of Premises}.}
    \label{tab:full_result_identification_of_premises}
\end{table*}

\begin{table*}[!htp]
\centering
\scriptsize
\begin{tabular}{l!{\color{gray!30}\vrule}llll!{\color{gray!30}\vrule}lll!{\color{gray!30}\vrule}ll}
\hline
\multirow{2}{*}{} &\multicolumn{4}{c}{Inputs} &\multicolumn{3}{c}{Automatic} &Semantic \\
\hline
&I &VP &CP &RS &BLEU-4 &ROUGE &CIDEr &BERT \\
\hline
\multirow{3}{*}{LLaMA3} & &\ding{51} & & &7.07 &28.41 &33.08 &43.00 \\
& &\ding{51} &\ding{51} & &8.65 ($\uparrow$ 1.58) &31.44 ($\uparrow$ 3.03) &40.87 ($\uparrow$ 7.79) &59.58 ($\uparrow$ 16.58) \\
& &\ding{51} &\ding{51} &\ding{51} &8.34 ($\downarrow$ 0.31) &31.18 ($\downarrow$ 0.26) &41.94 ($\uparrow$ 1.07) &56.70 ($\downarrow$ 2.88) \\
\hline
\multirow{3}{*}{Mistral} & &\ding{51} & & &2.95 &19.84 &23.28 &24.86 \\
& &\ding{51} &\ding{51} & &4.95 ($\uparrow$ 2.00) &25.13 ($\uparrow$ 5.29) &33.90 ($\uparrow$ 10.62) &39.92 ($\uparrow$ 16.05) \\
& &\ding{51} &\ding{51} &\ding{51} &6.15 ($\uparrow$ 1.20) &27.06 ($\uparrow$ 1.93) &38.34 ($\uparrow$ 4.43) &49.54 ($\uparrow$ 9.62) \\
\hline
\multirow{3}{*}{Zephyr} & &\ding{51} & & &2.78 &16.35 &24.06 &16.15 \\
& &\ding{51} &\ding{51} & &3.25 ($\uparrow$ 0.47) &17.44 ($\uparrow$ 1.08) &30.41 ($\uparrow$ 6.35) &31.38 ($\uparrow$ 6.52) \\
& &\ding{51} &\ding{51} &\ding{51} &5.20 ($\uparrow$ 1.94) &22.70 ($\uparrow$ 5.26) &36.29 ($\uparrow$ 5.88) &45.23 ($\uparrow$ 13.85) \\
\hline
\multirow{4}{*}{OFA} &\ding{51} & & & &0.00 &0.13 &0.01 &-41.26 \\
&\ding{51} &\ding{51} & & &0.00 (-) &5.24 ($\uparrow$ 5.10) &0.47 ($\uparrow$ 0.47) &-22.52 ($\uparrow$ 18.75) \\
&\ding{51} &\ding{51} &\ding{51} & &0.00 (-) &5.79 ($\uparrow$ 0.55) &0.37 ($\downarrow$ 0.10) &-15.87 ($\uparrow$ 6.65) \\
&\ding{51} &\ding{51} &\ding{51} &\ding{51} &0.00 (-) &6.53 ($\uparrow$ 0.75) &0.70 ($\uparrow$ 0.33) &-12.51 ($\uparrow$ 3.36) \\
\hline
\multirow{4}{*}{QwenVLChat} &\ding{51} & & & & 0.72 &13.12 &8.41 &14.32 \\
&\ding{51} &\ding{51} & & &4.02 ($\uparrow$ 3.30) &24.73 ($\uparrow$ 11.61) &30.58 ($\uparrow$ 22.17) &28.74 ($\uparrow$ 14.41) \\
&\ding{51} &\ding{51} &\ding{51} & &4.85 ($\uparrow$ 0.83) &26.67 ($\uparrow$ 1.94) &35.30 ($\uparrow$ 4.72) &34.05 ($\uparrow$ 5.31) \\
&\ding{51} &\ding{51} &\ding{51} &\ding{51} &4.89 ($\uparrow$ 0.03) &26.89 ($\uparrow$ 0.23) &38.30 ($\uparrow$ 3.00) &35.11 ($\uparrow$ 1.06) \\
\hline
\multirow{4}{*}{CogVLM} &\ding{51} & & & &4.96 &24.40 &25.56 &27.38 \\
&\ding{51} &\ding{51} & & &6.06 ($\uparrow$ 1.09) &27.37 ($\uparrow$ 2.97) &39.19 ($\uparrow$ 13.63) &33.24 ($\uparrow$ 5.87) \\
&\ding{51} &\ding{51} &\ding{51} & &7.18 ($\uparrow$ 1.13) &29.25 ($\uparrow$ 1.88) &47.21 ($\uparrow$ 8.02) &36.41 ($\uparrow$ 3.17) \\
&\ding{51} &\ding{51} &\ding{51} &\ding{51} &7.68 ($\uparrow$ 0.50) &30.03 ($\uparrow$ 0.78) &51.44 ($\uparrow$ 4.23) &37.71 ($\uparrow$ 1.29) \\
\hline
\multirow{4}{*}{Idefics2} &\ding{51} & & & &4.00 &21.97 &18.56 &21.27 \\
&\ding{51} &\ding{51} & & &4.53 ($\uparrow$ 0.53) &24.13 ($\uparrow$ 2.17) &28.79 ($\uparrow$ 10.24) &27.39 ($\uparrow$ 6.12) \\
&\ding{51} &\ding{51} &\ding{51} & &5.56 ($\uparrow$ 1.03) &25.17 ($\uparrow$ 1.04) &38.21 ($\uparrow$ 9.42) &33.22 ($\uparrow$ 5.83) \\
&\ding{51} &\ding{51} &\ding{51} &\ding{51} &7.48 ($\uparrow$ 1.92) &27.73 ($\uparrow$ 2.56) &53.22 ($\uparrow$ 15.01) &38.40 ($\uparrow$ 5.17) \\
\hline
\multirow{4}{*}{InstructBLIP} &\ding{51} & & & &0.00 &4.22 &1.01 &-15.92 \\
&\ding{51} &\ding{51} & & &3.27 ($\uparrow$ 3.26) &18.94 ($\uparrow$ 16.26) &22.16 ($\uparrow$ 21.15) &23.15 ($\uparrow$ 39.07) \\
&\ding{51} &\ding{51} &\ding{51} & &6.00 ($\uparrow$ 2.73) &26.91 ($\uparrow$ 7.32) &44.23 ($\uparrow$ 22.07) &35.20 ($\uparrow$ 12.05) \\
&\ding{51} &\ding{51} &\ding{51} &\ding{51} &6.27 ($\uparrow$ 0.27) &28.29 ($\uparrow$ 0.37) &45.53 ($\uparrow$ 1.29) &35.52 ($\uparrow$ 0.32) \\
\hline
\multirow{4}{*}{Unified-io 2} &\ding{51} & & & &0.07 &10.02 &0.89 &-8.49 \\
&\ding{51} &\ding{51} & & &0.65 ($\uparrow$ 0.58) &14.81 ($\uparrow$ 4.79) &5.18 ($\uparrow$ 4.29) &-0.04 ($\uparrow$ 8.45) \\
&\ding{51} &\ding{51} &\ding{51} & &0.82 ($\uparrow$ 0.17) &15.27 ($\uparrow$ 0.46) &7.73 ($\uparrow$ 2.55) &5.63 ($\uparrow$ 5.68) \\
&\ding{51} &\ding{51} &\ding{51} &\ding{51} &0.93 ($\uparrow$ 0.11) &16.10 ($\uparrow$ 0.83) &10.12 ($\uparrow$ 2.39) &8.43 ($\uparrow$ 2.79) \\
\hline
\multirow{4}{*}{LLaVA} &\ding{51} & & & &1.38 &14.93 &3.73 &3.25 \\
&\ding{51} &\ding{51} & & &3.92 ($\uparrow$ 2.54) &22.93 ($\uparrow$ 8.01) &21.49 ($\uparrow$ 17.76) &22.21 ($\uparrow$ 18.96) \\
&\ding{51} &\ding{51} &\ding{51} & &5.49 ($\uparrow$ 1.58) &26.40 ($\uparrow$ 3.47) &39.29 ($\uparrow$ 17.80) &32.48 ($\uparrow$ 10.28) \\
&\ding{51} &\ding{51} &\ding{51} &\ding{51} &5.68 ($\uparrow$ 0.19) &26.46 ($\uparrow$ 0.06) &42.65 ($\uparrow$ 3.36) &34.22 ($\uparrow$ 1.74) \\
\hline
\multirow{4}{*}{LLaVA-NeXT} &\ding{51} & & & &3.62 &21.08 &15.23 &18.05 \\
&\ding{51} &\ding{51} & & &6.78 ($\uparrow$ 3.16) &28.31 ($\uparrow$ 7.23) &42.17 ($\uparrow$ 26.94) &32.98 ($\uparrow$ 14.94) \\
&\ding{51} &\ding{51} &\ding{51} & &7.51 ($\uparrow$ 0.73) &30.03 ($\uparrow$ 1.72) &50.54 ($\uparrow$ 8.37) &37.93 ($\uparrow$ 4.95) \\
&\ding{51} &\ding{51} &\ding{51} &\ding{51} &8.51 ($\uparrow$ 1.00) &31.19 ($\uparrow$ 1.15) &61.70 ($\uparrow$ 11.16) &40.96 ($\uparrow$ 3.02) \\
\hline
\multirow{3}{*}{GPT-4-O} &\ding{51} & & & &2.38 &17.20 &23.08 &25.96 \\
&\ding{51} &\ding{51} & & &5.44 ($\uparrow$ 3.07) &24.47 ($\uparrow$ 7.27) &46.05 ($\uparrow$ 22.98) &36.09 ($\uparrow$ 10.13) \\
&\ding{51} &\ding{51} &\ding{51} &\ding{51} &6.82 ($\uparrow$ 1.38) &26.36 ($\uparrow$ 1.88) &61.20 ($\uparrow$ 15.15) &41.08 ($\uparrow$ 4.99) \\
\hline
\end{tabular}
\caption{Results on \textit{Deduction of Conclusion} in the Advertisement category.}
\label{tab:ads_conclusion_deduction}
\end{table*}
\begin{table*}[!htp]
\centering
\scriptsize
\begin{tabular}{l!{\color{gray!30}\vrule}llll!{\color{gray!30}\vrule}lll!{\color{gray!30}\vrule}ll}
\hline
\multirow{2}{*}{} &\multicolumn{4}{c}{Inputs} &\multicolumn{3}{c}{Automatic} &Semantic \\
\hline
&I &VP &CP &RS &BLEU-4 &ROUGE &CIDEr &BERT \\
\hline
\multirow{3}{*}{LLaMA3} & &\ding{51} & & &5.51 &27.67 &31.28 &26.46 \\
& &\ding{51} &\ding{51} & &6.65 ($\uparrow$ 1.13) &29.95 ($\uparrow$ 2.28) &48.01 ($\uparrow$ 16.73) &33.71 ($\uparrow$ 7.25) \\
& &\ding{51} &\ding{51} &\ding{51} &8.81 ($\uparrow$ 2.17) &32.17 ($\uparrow$ 2.22) &61.74 ($\uparrow$ 13.73) &39.20 ($\uparrow$ 5.49) \\
\hline
\multirow{3}{*}{Mistral} & &\ding{51} & & &1.87 &16.29 &11.24 &13.23 \\
& &\ding{51} &\ding{51} & &4.01 ($\uparrow$ 2.14) &22.24 ($\uparrow$ 5.95) &28.15 ($\uparrow$ 16.91) &25.22 ($\uparrow$ 11.99) \\
& &\ding{51} &\ding{51} &\ding{51} &6.45 ($\uparrow$ 2.44) &27.82 ($\uparrow$ 5.59) &45.00 ($\uparrow$ 16.85) &34.39 ($\uparrow$ 9.17) \\
\hline
\multirow{3}{*}{Zephyr} & &\ding{51} & & &1.70 &13.88 &11.54 &16.15 \\
& &\ding{51} &\ding{51} & &3.28 ($\uparrow$ 1.58) &17.51 ($\uparrow$ 3.63) &24.92 ($\uparrow$ 13.38) &26.40 ($\uparrow$ 10.25) \\
& &\ding{51} &\ding{51} &\ding{51} &6.91 ($\uparrow$ 3.63) &27.20 ($\uparrow$ 9.69) &46.63 ($\uparrow$ 21.71) &36.70 ($\uparrow$ 10.29) \\
\hline
\multirow{4}{*}{OFA} &\ding{51} & & & &0.00 &0.45 &0.01 &-41.35 \\
&\ding{51} &\ding{51} & & &0.00 (-) &5.30 ($\uparrow$ 4.84) &0.27 ($\uparrow$ 0.26) &-27.23 ($\uparrow$ 14.12) \\
&\ding{51} &\ding{51} &\ding{51} & &0.00 (-) &8.15 ($\uparrow$ 2.85) &0.26 ($\downarrow$ 0.01) &-17.30 ($\uparrow$ 9.93) \\
&\ding{51} &\ding{51} &\ding{51} &\ding{51} &0.00 (-) &8.45 ($\uparrow$ 0.30) &0.58 ($\uparrow$ 0.32) &-15.74 ($\uparrow$ 1.56) \\
\hline
\multirow{4}{*}{QwenVLChat} &\ding{51} & & & &0.49 &13.98 &4.23 &10.79 \\
&\ding{51} &\ding{51} & & &3.18 ($\uparrow$ 2.69) &23.80 ($\uparrow$ 9.81) &18.02 ($\uparrow$ 13.79) &17.18 ($\uparrow$ 6.39) \\
&\ding{51} &\ding{51} &\ding{51} & &4.23 ($\uparrow$ 1.05) &26.40 ($\uparrow$ 2.60) &26.76 ($\uparrow$ 8.74) &25.03 ($\uparrow$ 7.85) \\
&\ding{51} &\ding{51} &\ding{51} &\ding{51} &4.79 ($\uparrow$ 0.56) &27.87 ($\uparrow$ 1.47) &32.10 ($\uparrow$ 5.34) &29.49 ($\uparrow$ 4.46) \\
\hline
\multirow{4}{*}{CogVLM} &\ding{51} & & & &4.89 &27.48 &24.56 &23.51 \\
&\ding{51} &\ding{51} & & &5.49 ($\uparrow$ 0.60) &28.50 ($\uparrow$ 1.02) &32.11 ($\uparrow$ 7.54) &27.24 ($\uparrow$ 3.73) \\
&\ding{51} &\ding{51} &\ding{51} & &6.43 ($\uparrow$ 0.94) &29.64 ($\uparrow$ 1.14) &38.02 ($\uparrow$ 5.91) &29.88 ($\uparrow$ 2.64) \\
&\ding{51} &\ding{51} &\ding{51} &\ding{51} &7.89 ($\uparrow$ 1.46) &31.72 ($\uparrow$ 2.08) &53.05 ($\uparrow$ 15.04) &34.45 ($\uparrow$ 4.56) \\
\hline
\multirow{4}{*}{Idefics2} &\ding{51} & & & &3.50 &21.74 &16.07 &16.36 \\
&\ding{51} &\ding{51} & & &3.61 ($\uparrow$ 0.78) &23.47 ($\uparrow$ 2.04) &20.02 ($\uparrow$ 7.20) &16.70 ($\uparrow$ 6.78) \\
&\ding{51} &\ding{51} &\ding{51} & &5.42 ($\uparrow$ 1.81) &26.58 ($\uparrow$ 3.11) &29.96 ($\uparrow$ 9.94) &24.50 ($\uparrow$ 7.81) \\
&\ding{51} &\ding{51} &\ding{51} &\ding{51} &8.28 ($\uparrow$ 2.86) &31.01 ($\uparrow$ 4.43) &54.64 ($\uparrow$ 24.68) &34.27 ($\uparrow$ 9.76) \\
\hline
\multirow{4}{*}{InstructBLIP} &\ding{51} & & & &0.00 &3.24 &0.40 &-21.55 \\
&\ding{51} &\ding{51} & & &2.53 ($\uparrow$ 2.53) &18.94 ($\uparrow$ 15.14) &16.35 ($\uparrow$ 15.60) &16.62 ($\uparrow$ 34.98) \\
&\ding{51} &\ding{51} &\ding{51} & &5.33 ($\uparrow$ 2.80) &26.91 ($\uparrow$ 7.98) &36.87 ($\uparrow$ 20.52) &28.92 ($\uparrow$ 12.30) \\
&\ding{51} &\ding{51} &\ding{51} &\ding{51} &6.18 ($\uparrow$ 0.85) &28.29 ($\uparrow$ 1.38) &42.53 ($\uparrow$ 5.65) &32.17 ($\uparrow$ 3.25) \\
\hline
\multirow{4}{*}{Unified-io 2} &\ding{51} & & & &0.00 &9.07 &0.40 &-11.68 \\
&\ding{51} &\ding{51} & & &0.56 ($\uparrow$ 0.56) &11.32 ($\uparrow$ 2.25) &2.60 ($\uparrow$ 2.20) &-7.80 ($\uparrow$ 3.88) \\
&\ding{51} &\ding{51} &\ding{51} & &0.62 ($\uparrow$ 0.06) &13.79 ($\uparrow$ 2.48) &5.40 ($\uparrow$ 2.80) &2.39 ($\uparrow$ 10.19) \\
&\ding{51} &\ding{51} &\ding{51} &\ding{51} &1.33 ($\uparrow$ 0.71) &16.50 ($\uparrow$ 2.70) &12.63 ($\uparrow$ 7.24) &7.47 ($\uparrow$ 5.08) \\
\hline
\multirow{4}{*}{LLaVA} &\ding{51} & & & &1.46 &17.84 &3.55 &0.92 \\
&\ding{51} &\ding{51} & & &3.73 ($\uparrow$ 2.27) &23.22 ($\uparrow$ 5.38) &12.56 ($\uparrow$ 9.01) &14.92 ($\uparrow$ 14.01) \\
&\ding{51} &\ding{51} &\ding{51} & &5.61 ($\uparrow$ 1.87) &27.63 ($\uparrow$ 4.41) &28.99 ($\uparrow$ 16.44) &25.89 ($\uparrow$ 10.97) \\
&\ding{51} &\ding{51} &\ding{51} &\ding{51} &7.87 ($\uparrow$ 2.26) &30.46 ($\uparrow$ 2.84) &47.25 ($\uparrow$ 18.25) &33.11 ($\uparrow$ 7.22) \\
\hline
\multirow{4}{*}{LLaVA-NeXT} &\ding{51} & & & &2.71 &22.32 &11.26 &11.26 \\
&\ding{51} &\ding{51} & & &5.78 ($\uparrow$ 3.07) &27.79 ($\uparrow$ 5.48) &28.12 ($\uparrow$ 16.86) &22.46 ($\uparrow$ 11.20) \\
&\ding{51} &\ding{51} &\ding{51} & &7.31 ($\uparrow$ 1.52) &30.44 ($\uparrow$ 2.65) &43.84 ($\uparrow$ 15.73) &29.57 ($\uparrow$ 7.11) \\
&\ding{51} &\ding{51} &\ding{51} &\ding{51} &9.16 ($\uparrow$ 1.85) &33.09 ($\uparrow$ 2.65) &61.44 ($\uparrow$ 17.60) &35.88 ($\uparrow$ 6.32) \\
\hline
\multirow{3}{*}{GPT-4-O} &\ding{51} & & & &4.07 &23.37 &23.15 &24.96 \\
&\ding{51} &\ding{51} & & &6.40 ($\uparrow$ 2.34) &27.39 ($\uparrow$ 4.02) &40.64 ($\uparrow$ 17.50) &32.05 ($\uparrow$ 7.09) \\
&\ding{51} &\ding{51} &\ding{51} &\ding{51} &8.69 ($\uparrow$ 2.29) &31.32 ($\uparrow$ 3.93) &63.13 ($\uparrow$ 22.48) &40.78 ($\uparrow$ 8.73) \\
\hline

\end{tabular}
\caption{Results on \textit{Deduction of Conclusion} in the Cartoon category.}
\label{tab:cartoon_conclusion_deduction}
\end{table*}
\begin{table*}[!htp]
\centering
\scriptsize
\begin{tabular}{l!{\color{gray!30}\vrule}llll!{\color{gray!30}\vrule}lll!{\color{gray!30}\vrule}ll}
\hline
\multirow{2}{*}{} &\multicolumn{4}{c}{Inputs} &\multicolumn{3}{c}{Automatic} &Semantic \\
\hline
&I &VP &CP &RS &BLEU-4 &ROUGE &CIDEr &BERT \\
\hline
\multirow{3}{*}{LLaMA3} & &\ding{51} & & &6.40 &28.09 &37.93 &30.22 \\
& &\ding{51} &\ding{51} & &7.78 ($\uparrow$ 1.39) &30.80 ($\uparrow$ 2.71) &54.57 ($\uparrow$ 16.65) &37.77 ($\uparrow$ 7.55) \\
& &\ding{51} &\ding{51} &\ding{51} &8.54 ($\uparrow$ 0.76) &31.61 ($\uparrow$ 0.82) &58.88 ($\uparrow$ 4.31) &40.75 ($\uparrow$ 2.98) \\
\hline
\multirow{3}{*}{Mistral} & &\ding{51} & & &2.48 &18.30 &18.41 &18.93 \\
& &\ding{51} &\ding{51} & &4.54 ($\uparrow$ 2.06) &23.88 ($\uparrow$ 5.57) &34.83 ($\uparrow$ 16.42) &30.15 ($\uparrow$ 11.21) \\
& &\ding{51} &\ding{51} &\ding{51} &6.28 ($\uparrow$ 1.74) &27.39 ($\uparrow$ 3.51) &47.58 ($\uparrow$ 12.75) &36.63 ($\uparrow$ 6.48) \\
\hline
\multirow{3}{*}{Zephyr} & &\ding{51} & & &2.31 &15.28 &19.10 &20.64 \\
& &\ding{51} &\ding{51} & &3.26 ($\uparrow$ 0.95) &17.47 ($\uparrow$ 2.18) &28.59 ($\uparrow$ 9.49) &28.67 ($\uparrow$ 8.04) \\
& &\ding{51} &\ding{51} &\ding{51} &5.94 ($\uparrow$ 2.67) &24.65 ($\uparrow$ 7.18) &45.84 ($\uparrow$ 17.25) &36.47 ($\uparrow$ 7.79) \\
\hline
\multirow{4}{*}{OFA} &\ding{51} & & & &0.00 &0.27 &0.01 &-41.30 \\
&\ding{51} &\ding{51} & & &0.00 (-) &5.26 ($\uparrow$ 4.99) &0.39 ($\uparrow$ 0.38) &-24.55 ($\uparrow$ 5.68) \\
&\ding{51} &\ding{51} &\ding{51} & &0.00 (-) &6.81 ($\uparrow$ 1.55) &0.32 ($\downarrow$ 0.06) &-16.49 ($\uparrow$ 1.73) \\
&\ding{51} &\ding{51} &\ding{51} &\ding{51} &0.00 (-) &7.36 ($\uparrow$ 0.55) &0.65 ($\uparrow$ 0.32) &-13.91 ($\uparrow$ 1.22) \\
\hline
\multirow{4}{*}{QwenVLChat} &\ding{51} & & & &0.62 &13.50 &6.60 &12.79 \\
&\ding{51} &\ding{51} & & &3.66 ($\uparrow$ 3.04) &24.33 ($\uparrow$ 10.83) &25.15 ($\uparrow$ 18.54) &23.74 ($\uparrow$ 10.94) \\
&\ding{51} &\ding{51} &\ding{51} & &4.58 ($\uparrow$ 0.92) &26.55 ($\uparrow$ 2.23) &31.61 ($\uparrow$ 6.46) &30.15 ($\uparrow$ 6.41) \\
&\ding{51} &\ding{51} &\ding{51} &\ding{51} &4.85 ($\uparrow$ 0.26) &27.31 ($\uparrow$ 0.76) &35.62 ($\uparrow$ 4.01) &32.68 ($\uparrow$ 2.53) \\
\hline
\multirow{4}{*}{Idefics2} &\ding{51} & & & &3.50 &21.74 &16.07 &16.36 \\
&\ding{51} &\ding{51} & & &4.13 ($\uparrow$ 0.64) &23.84 ($\uparrow$ 2.11) &25.00 ($\uparrow$ 8.92) &22.76 ($\uparrow$ 6.41) \\
&\ding{51} &\ding{51} &\ding{51} & &5.50 ($\uparrow$ 1.37) &25.78 ($\uparrow$ 1.93) &34.64 ($\uparrow$ 9.64) &29.45 ($\uparrow$ 6.69) \\
&\ding{51} &\ding{51} &\ding{51} &\ding{51} &7.82 ($\uparrow$ 2.32) &29.15 ($\uparrow$ 3.37) &53.84 ($\uparrow$ 19.20) &36.61 ($\uparrow$ 7.16) \\
\hline
\multirow{4}{*}{InstructBLIP} &\ding{51} & & & &0.00 &3.80 &0.75 &-18.36 \\
&\ding{51} &\ding{51} & & &2.53 ($\uparrow$ 2.53) &18.94 ($\uparrow$ 15.14) &16.35 ($\uparrow$ 15.60) &16.62 ($\uparrow$ 34.98) \\
&\ding{51} &\ding{51} &\ding{51} & &5.33 ($\uparrow$ 2.80) &26.91 ($\uparrow$ 7.98) &36.87 ($\uparrow$ 20.52) &28.92 ($\uparrow$ 12.30) \\
&\ding{51} &\ding{51} &\ding{51} &\ding{51} &6.18 ($\uparrow$ 0.85) &28.29 ($\uparrow$ 1.38) &42.53 ($\uparrow$ 5.65) &32.17 ($\uparrow$ 3.25) \\
\hline
\multirow{4}{*}{CogVLM} &\ding{51} & & & &4.93 &25.73 &25.13 &25.53 \\
&\ding{51} &\ding{51} & & &5.81 ($\uparrow$ 0.88) &27.86 ($\uparrow$ 2.13) &36.13 ($\uparrow$ 11.00) &30.65 ($\uparrow$ 4.94) \\
&\ding{51} &\ding{51} &\ding{51} & &6.86 ($\uparrow$ 1.04) &29.42 ($\uparrow$ 1.56) &43.23 ($\uparrow$ 7.10) &33.59 ($\uparrow$ 2.94) \\
&\ding{51} &\ding{51} &\ding{51} &\ding{51} &7.77 ($\uparrow$ 0.92) &30.76 ($\uparrow$ 1.34) &52.14 ($\uparrow$ 8.90) &36.30 ($\uparrow$ 2.71) \\
\hline
\multirow{4}{*}{Unified-io 2} &\ding{51} & & & &0.04 &9.61 &0.68 &-9.87 \\
&\ding{51} &\ding{51} & & &0.61 ($\uparrow$ 0.57) &13.30 ($\uparrow$ 3.69) &4.07 ($\uparrow$ 3.39) &-3.40 ($\uparrow$ 6.47) \\
&\ding{51} &\ding{51} &\ding{51} & &0.74 ($\uparrow$ 0.12) &14.63 ($\uparrow$ 1.33) &6.72 ($\uparrow$ 2.66) &4.23 ($\uparrow$ 7.63) \\
&\ding{51} &\ding{51} &\ding{51} &\ding{51} &1.10 ($\uparrow$ 0.37) &16.27 ($\uparrow$ 1.64) &11.21 ($\uparrow$ 4.48) &8.01 ($\uparrow$ 3.78) \\
\hline
\multirow{4}{*}{LLaVA} &\ding{51} & & & &1.50 &16.01 &3.65 &2.24 \\
&\ding{51} &\ding{51} & & &3.86 ($\uparrow$ 2.36) &22.88 ($\uparrow$ 6.87) &18.69 ($\uparrow$ 15.04) &19.98 ($\uparrow$ 17.74) \\
&\ding{51} &\ding{51} &\ding{51} & &5.54 ($\uparrow$ 1.69) &26.93 ($\uparrow$ 4.05) &34.84 ($\uparrow$ 16.14) &29.63 ($\uparrow$ 9.66) \\
&\ding{51} &\ding{51} &\ding{51} &\ding{51} &6.63 ($\uparrow$ 1.09) &28.19 ($\uparrow$ 1.26) &44.64 ($\uparrow$ 9.80) &33.74 ($\uparrow$ 4.11) \\
\hline
\multirow{4}{*}{LLaVA-NeXT} &\ding{51} & & & &3.23 &21.62 &13.51 &15.11 \\
&\ding{51} &\ding{51} & & &6.35 ($\uparrow$ 3.12) &28.09 ($\uparrow$ 6.47) &36.09 ($\uparrow$ 22.58) &28.43 ($\uparrow$ 16.75) \\
&\ding{51} &\ding{51} &\ding{51} & &7.42 ($\uparrow$ 1.07) &30.21 ($\uparrow$ 2.12) &47.64 ($\uparrow$ 11.55) &34.31 ($\uparrow$ 8.07) \\
&\ding{51} &\ding{51} &\ding{51} &\ding{51} &8.46 ($\uparrow$ 1.04) &31.69 ($\uparrow$ 1.49) &61.14 ($\uparrow$ 13.50) &39.50 ($\uparrow$ 2.58) \\
\hline
\multirow{3}{*}{GPT-4-O} &\ding{51} & & & &3.11 &19.87 &23.11 &24.50 \\
&\ding{51} &\ding{51} & & &5.86 ($\uparrow$  2.75) &25.74 ($\uparrow$ 5.87) &43.71 ($\uparrow$ 20.61) &31.89 ($\uparrow$ 7.39) \\
&\ding{51} &\ding{51} &\ding{51} &\ding{51} &7.63 ($\uparrow$  1.77) &28.51 ($\uparrow$ 2.77) &62.03 ($\uparrow$ 18.32) &38.42 ($\uparrow$ 6.53) \\
\hline
\end{tabular}
\caption{Results for the \textit{Deduction of Conclusion} averaged across the two categories.}
\label{tab:total_conclusion_deduction}
\end{table*}
\newpage


\end{document}